\documentclass[11pt,a4paper]{article}
\usepackage{times,latexsym}
\usepackage{url}
\usepackage[T1]{fontenc}
\usepackage{graphicx}
\usepackage{multicol, blindtext}
\usepackage{CJK} 
\usepackage{times}
\usepackage{latexsym}
\usepackage{bbm}
\usepackage{array}
\usepackage{subfigure}
\usepackage{multirow}
\usepackage{amsthm}
\usepackage{amsmath}
\usepackage{amssymb}
\usepackage{amsthm}
\usepackage{color}
\usepackage{epsfig}
\usepackage{eqnarray,amsmath}
\usepackage{amsfonts}
\usepackage{txfonts}
\usepackage{algorithmicx}
\usepackage{times}
\usepackage[linesnumbered,ruled,commentsnumbered]{algorithm2e}
\newtheorem{myDef}{Definition} 

\usepackage[hyperref]{acl2019}
\aclfinalcopy

\usepackage{xspace,mfirstuc,tabulary}

\newif\iftaclinstructions
\taclinstructionsfalse 
\iftaclinstructions
\renewcommand{\confidential}{}
\renewcommand{\anonsubtext}{(No author info supplied here, for consistency with
TACL-submission anonymization requirements)}
\newcommand{\instr}
\fi

\title{DuTongChuan\thanks{DuTongChuan, abbreviation of Baidu simultaneous interpreting in Chinese Pinyin.} : Context-aware Translation Model for Simultaneous Interpreting }

\author{
 Hao Xiong, Ruiqing Zhang, Chuanqiang Zhang, Zhongjun He\\
  \textbf{Hua Wu} and \textbf{Haifeng Wang} \\
 Baidu Inc. No. 10, Shangdi 10th Street \\
 Beijing, 100085, China \\
  {\sf \{xionghao05, zhangruiqing01, zhangchuanqiang\}} \\ \{hezhongjun, wu\_hua, wanghaifeng\}@baidu.com}

\date{}

\begin{document}
\begin{CJK*}{UTF8}{gbsn}
\maketitle
\begin{abstract}
  In this paper, we present \textbf{DuTongChuan}, a novel context-aware translation model for simultaneous interpreting. This model allows to constantly read streaming text from the Automatic Speech Recognition (ASR) model and simultaneously determine the boundaries of {\em Information Units} (IUs) one after another. 
 The detected IU is then translated into a fluent translation with two simple yet effective decoding strategies: partial decoding and context-aware decoding. 
 In practice, by controlling the granularity of IUs and the size of the context, we can get a good trade-off between latency and translation quality easily. 
Elaborate evaluation from human translators reveals that our system achieves promising translation quality (85.71\% for Chinese-English, and 86.36\% for English-Chinese), specially in the sense of surprisingly good discourse coherence. According to an End-to-End (speech-to-speech simultaneous interpreting) evaluation, this model presents impressive performance in reducing latency (to less than 3 seconds at most times). Furthermore, we successfully deploy this model in a variety of Baidu's products which have hundreds of millions of users, and we release it as a service in our AI platform \footnote{\url{https://fanyi-api.baidu.com/api/trans/product/simultaneous}}. 
\end{abstract}

\section{Introduction}
\label{intro}
Recent progress in Automatic Speech Recognition (ASR) and Neural Machine Translation (NMT), has facilitated the research on automatic speech translation with applications to live and streaming scenarios such as Simultaneous Interpreting (SI). 
In contrast to non-real time speech translation, simultaneous interpreting involves starting translating source speech, before the speaker finishes speaking (translating the on-going speech while listening to it). Because of this distinguishing feature, simultaneous interpreting is widely used by multilateral organizations (UN/EU), international summits (APEC/G-20), legal proceedings, and press conferences. Despite of recent advance \cite{DBLP:journals/corr/abs-1810-08398,arivazhagan2019monotonic}, the research on simultaneous interpreting is notoriously difficult \cite{DBLP:journals/corr/abs-1810-08398} due to well known challenging requirements: high-quality translation and low latency.  
\begin{figure}[t]
\centering
\includegraphics[width=\linewidth]{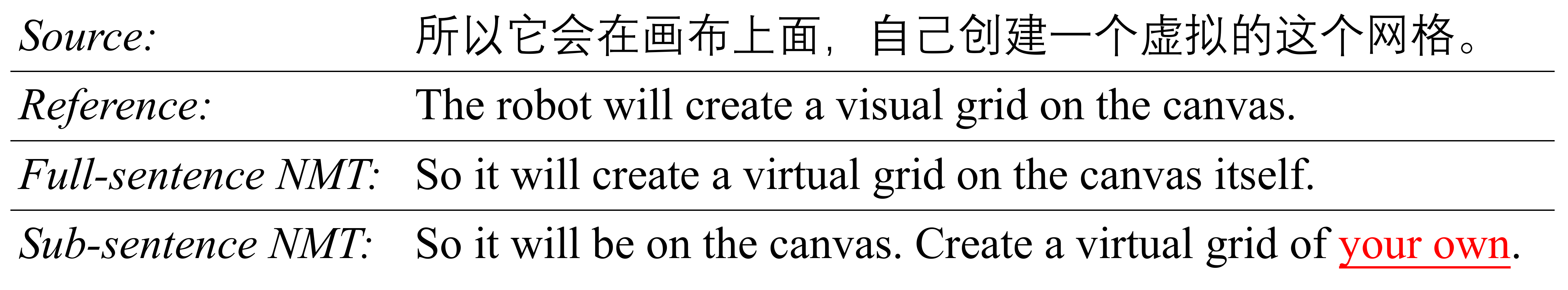} 
\caption{For this sentence, a full-sentence NMT model produces an appropriate translation with, however, a long latency in the context of simultaneous translation, as it needs to wait until the end of the full sentence to start translating. In contrast, a sub-sentence NMT model outputs a translation with less coherence and fluency, although it has a relatively short latency as it starts translating after reading the comma in the source text. }
\label{fig:discourse}
\end{figure}

\begin{figure*}[t]
\centering
\includegraphics[width=\linewidth]{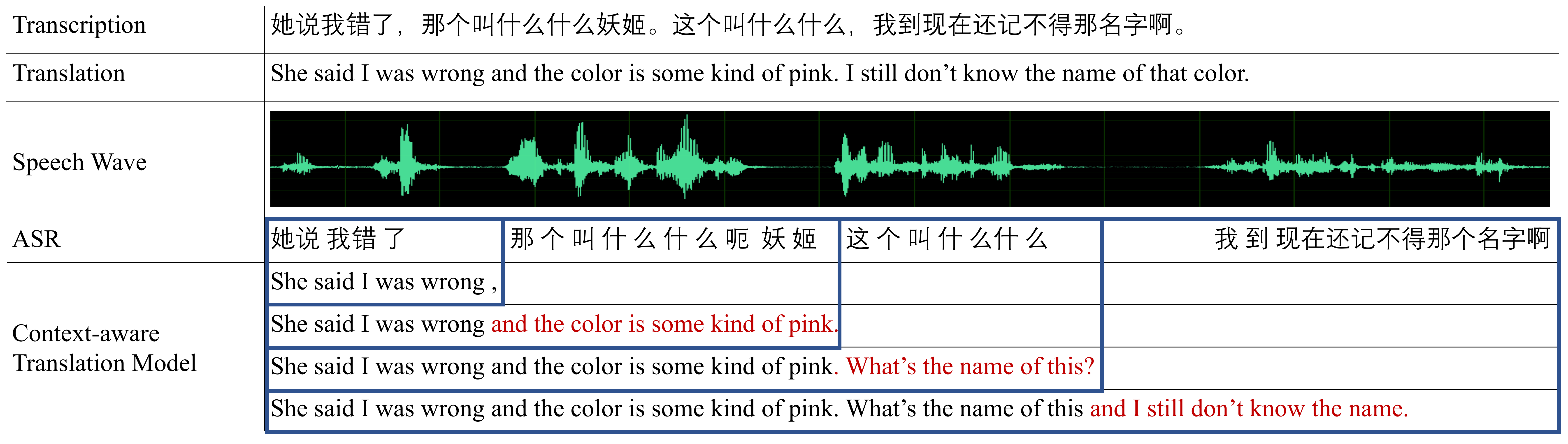}
\caption{This example shows a special case using sub-sentences as our information units. The blue solid squares indicate the scope of source context we use for translation. The text in red are incrementally generated translations. We discard a preceding generated token to make a coherent translation. }
\label{fig:intro}
\end{figure*}

Many studies present methods to improve the translation quality by enhancing the robustness of translation model against ASR errors \cite{tsvetkov2014augmenting,chen2017mitigating,sperber2017toward,cheng2018towards,liu2018robust,li2018improving}. On the other hand, to reduce latency, some researchers propose models that start translating after reading a few source tokens \cite{fujita2013simple,grissom2014don,cho2016can,gu2017learning,niehues2018,arivazhagan2019monotonic}. As one representative work related to this topic, recently, we present a translation model using prefix-to-prefix framework with $wait-k$ policy \cite{DBLP:journals/corr/abs-1810-08398}. This model is simple yet effective in practice, achieving impressive performance both on translation quality and latency.

However, existing work pays less attention to the fluency of translation, which is extremely important in the context of simultaneous translation. For example, we have a sub-sentence NMT model that starts to translate after reading a sub-sentence rather than waiting until the end of a sentence like the full-sentence models does. This will definitely reduce the time waiting for the source language speech. However, as shown in the Figure \ref{fig:discourse}, the translation for each sub-sentence is barely adequate, whereas the translation of the entire source sentence lacks coherence and fluency. Moreover, it is clear that the model produces an inappropriate translation ``your own'' for the source token ``自己'' due to the absence of the preceding sub-sentence. 

To make the simultaneous machine translation more accessible and producible, we borrow SI strategies used by human interpreters to create our model. 
As shown in Figure \ref{fig:intro}, this model is able to constantly read streaming text from the ASR model, and simultaneously determine the boundaries of Information Units (IUs) one after another. Each detected IU is then translated into a fluent translation with two simple yet effective decoding strategies: partial decoding and context-aware decoding. Specifically, IUs at the beginning of each sentence are sent to the partial decoding module. Other information units, either appearing in the middle or at the end of a sentence, are translated into target language by the context-aware decoding module. Notice that this module is able to exploit additional context from the history so that the model can generate coherent translation.  
This method is derived from the ``salami technique'' \cite{roderick1998conference,gile2009basic}, or ``chunking'', one of the most commonly used strategies by human interpreters to cope with the linearity constraint in simultaneous interpreting. Having severely limited access to source speech structure in SI, interpreters tend to slice up the incoming speech into smaller meaningful pieces that can be directly rendered or locally reformulated without having to wait for the entire sentence to unfold.

In general, there are several remarkable novel advantages that differ our model from the previous work: 
\begin{itemize}
\item We propose a practical solution for simultaneous interpreting including an information unit detector and a tailored NMT model.
\item We can trade off latency and translation quality easily by controlling the granularity of IUs and the size of the context.
\item The mechanism of context-aware decoding module make our model generate fluent translation. 
\end{itemize}
 \begin{figure*}[t]
\centering
\includegraphics[width=\linewidth]{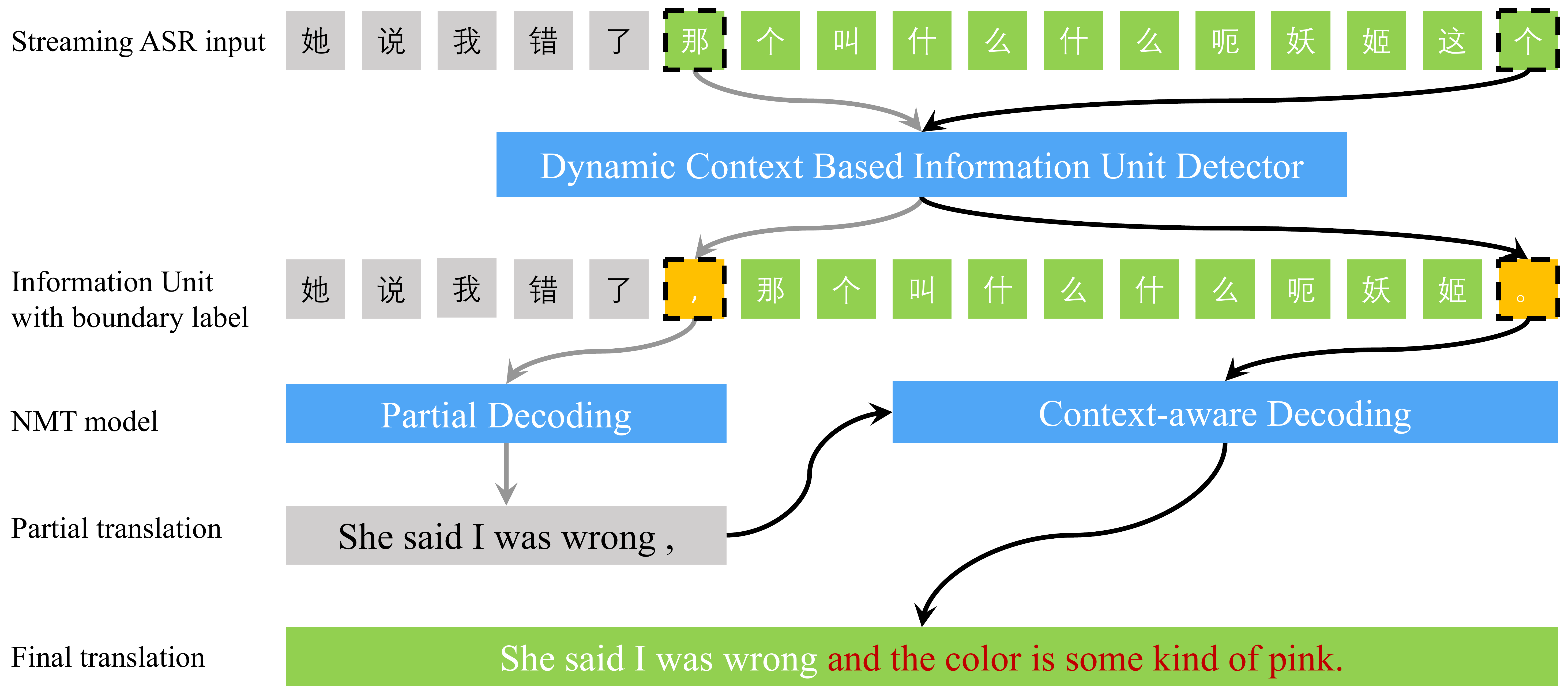}
\caption{In our context-aware translation model, the boundaries of information units in streaming ASR input are determined by a novel IU boundary detector. IUs at different positions are translated using different NMT models, if an IU stands at the beginning of a sentence, then it will be translated by the partial decoding module. Otherwise, context-aware decoding is applied to translate the IU into a coherent translation. Notice that the dashed squares in the first line denote the anchor to determine the IU boundary. }
\label{fig:cnmt}
\end{figure*}

For a comprehensive evaluation of our system, we use two evaluation metrics: translation quality and latency. 
 According to the automatic evaluation metric, our system presents excellent performance both in translation quality and latency. In the speech-to-speech scenario, our model achieves an acceptability of 85.71\% for Chinese-English translation, and 86.36\% for English-Chinese translation in human evaluation. Moreover, the output speech lags behind the source speech by an average of less than 3 seconds, which presents surprisingly good experience for machine translation users \cite{lee2002ear,lamberger2001text,timarova2015simultaneous}. We also ask three interpreters with SI experience to simultaneously interpret the test speech in a mock conference setting. However, the target texts transcribed from human SI obtain worse BLEU scores as the reference in the test set are actually from written translating rather than simultaneous interpreting. More importantly, when evaluated by human translators, the performance of NMT model is comparable to the professional human interpreter.

 The contributions of this paper can be concluded into the following aspects:
 \begin{itemize}
 \item We propose a novel context-aware translation model for simultaneous interpreting.
 \item We deliver a novel speech translation corpus for evaluating simultaneous machine translation.
 \item We conduct elaborate experiments showing our context-aware translation model's impressive performance in improving translation quality and shortening latency.
 \item We propose a novel comparison between the text results from human simultaneous interpreting and machine translation.
 \end{itemize}
\section{Context-aware Translation Model}
As shown in Figure \ref{fig:cnmt}, our model consists of two key modules: an information unit boundary detector and a tailored NMT model. 
In the process of translation, the IU detector will determine the boundary for each IU while constantly reading the steaming input from the ASR model.
Then, different decoding strategies are applied to translate IUs at the different positions.

In this section, we use ``IU'' to denote one \textbf{sub-sentence} for better description. But in effect, our translation model is a general solution for simultaneous interpreting, and is compatible to IUs at arbitrary granularity, i.e., clause-level, phrase-level, and word-level, etc. 

For example, by treating a full-sentence as an IU, the model is reduced to the standard translation model. When the IU is one segment, it is reduced to the \textit{segment-to-segment} translation model \cite{oda2014optimizing,niehues2018}. Moreover, if we treat one token as an IU, it is reduced to our previous \textit{wait-k} model \cite{DBLP:journals/corr/abs-1810-08398}. The key point of our model is to train the IU detector to recognize the IU boundary at the corresponding granularity. 

In the remain of this section, we will introduce above two components in details.
\subsection{Dynamic Context Based Information Unit Boundary Detector}

 \begin{figure*}[t]
\centering
\includegraphics[width=\linewidth]{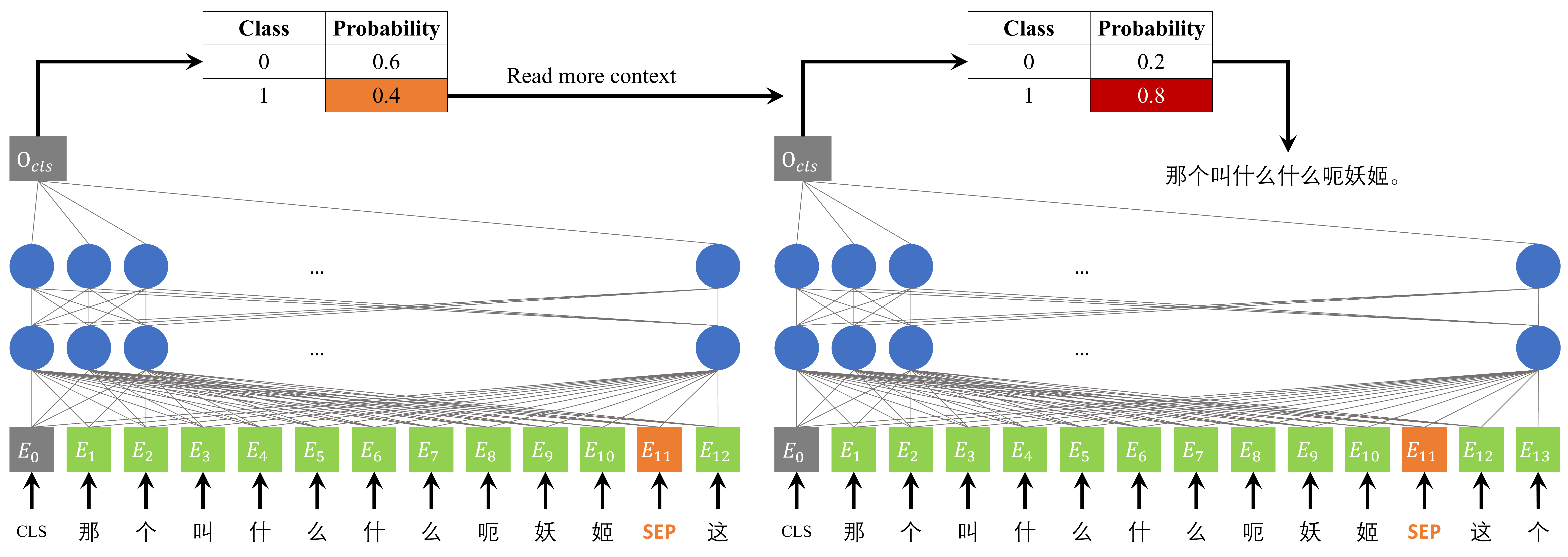}
\caption{A running example of our dynamic context based IU boundary detector. In this example, the model learns to determine the classification of the current {\em anchor}, ``姬'' (we insert an additional symbol, \textbf{SEP} to be consistent with the training format in the work of \newcite{devlin2019bert}). If the probability (0.4 in left side case) of decision for a boundary at the present \textit{anchor} is smaller than a threshold, i.e., $0.7$, then it is necessary to consider more context (additional context: ``这个'') to make a reliable decision (0.8 in right side case).   }
\label{fig:sbd}
\end{figure*}
Recent success on pre-training indicates that a pre-trained language representation is beneficial to downstream natural language processing tasks including classification and sequence labeling problems \cite{devlin2019bert,sun2019ernie,yang2019xlnet}. We thus formulate the IU boundary detection as a classification problem, and fine-tune the pre-trained model on a small size training corpus. Fine-tuned in several iterations, the model learns to recognize the boundaries of information units correctly.

As shown in Figure \ref{fig:sbd}, the model tries to predict the potential class for the current position. Once the position is assigned to a definitely positive class, its preceding sequence is labeled as one information unit. One distinguishing feature of this model is that we allow it to wait for more context so that it can make a reliable prediction. We call this model a dynamic context based information unit boundary detector. 

 \begin{myDef}
 Assuming the model has already read a sequence $(x_1,x_2,...,x_t, ..., x_n)$ with $n$ tokens, we denote $x_t$ as the \textbf{anchor}, and the subsequence $(x_{t+1}, ..., x_n)$ with $n-t$ tokens as
  \textbf{dynamic context}.  \label{def:1}
 \end{myDef}
For example, in Figure \ref{fig:sbd}, the \textbf{anchor} in both cases is ``姬'', and the \textbf{dynamic context} in the left side case is ``这'', and in the right side case is ``这个''.
 \begin{myDef} If the normalized probability $p_{x_t}=p(c=1|x_1,...,x_n,\theta)$ for the prediction of the current \textbf{anchor} $x_t$ is larger than a threshold $\delta_1$, then the sequence $(x_1,x_2,...,x_t)$ is a \textbf{complete sequence}, and if $p_{x_t}$ is smaller than a threshold $\delta_2$ ($\delta_2 < \delta_1$), it is an \textbf{incomplete sequence}, otherwise it is an \textbf{undetermined sequence}.
 \label{def:2} 
   \end{myDef}
 For a complete sequence $(x_1,x_2,...,x_t)$, we will send it to the corresponding translation model \footnote{We can develop an additional model to predict the punctuation for the complete sequence. It is also available to extend the detector to predict the punctuation directly, i.g., 0:no punctuation; 1:comma; 2:period; 3:question mark, etc.}. Afterwards, the detector will continue to recognize boundaries in the rest of the sequence ($x_{t+1},x_{t+2},...,x_n$). For an incomplete sequence, we will take the $x_{t+1}$ as the new anchor for further detection. For an undetermined sequence, which is as shown in Figure \ref{fig:sbd}, the model will wait for a new token $x_{n+1}$, and take ($x_{t+1},...,x_{n+1}$) as dynamic context for further prediction.

 \begin{figure*}[t!]
\centering
\includegraphics[width=\linewidth]{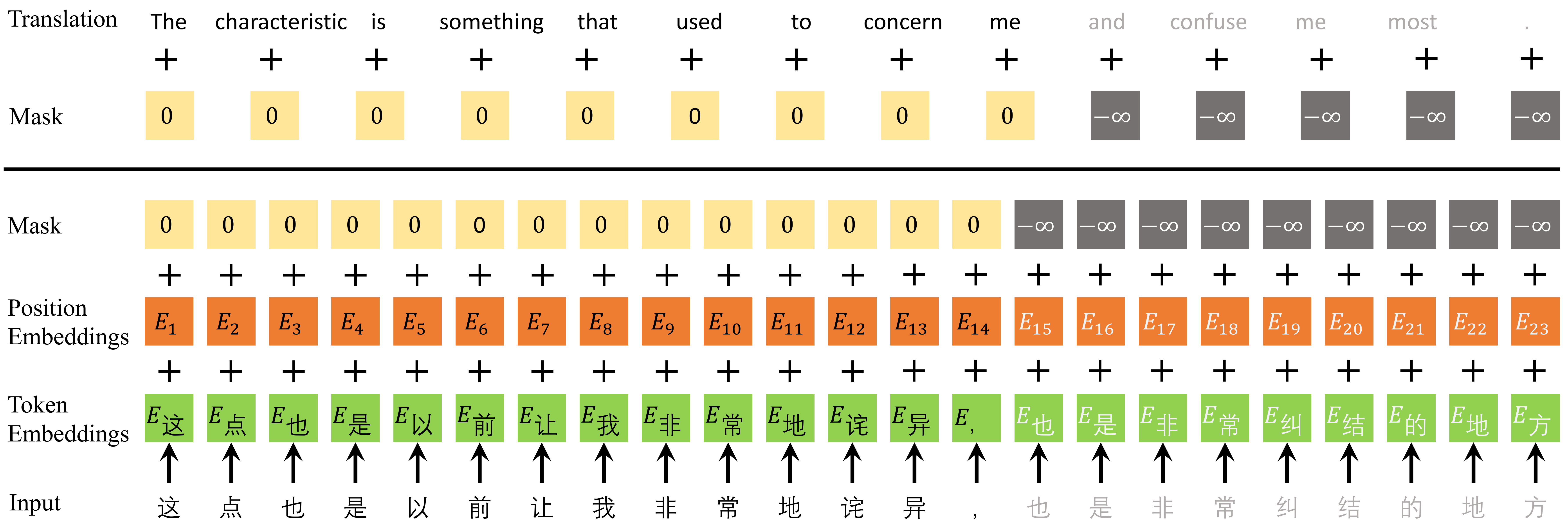}
\caption{Source and target representation for training partial decoding model, where we mask the second sub-sentence by summing a negative infinite number when training the partial decoding model. For simplicity, we omit the embeddings for the target side.     }
\label{fig:partial}
\end{figure*}
In the training stage, for one common sentence including two sub-sequences, $s_1$ and $s_2$.
 We collect $(s_1,\textit{SEP})$ plus any token in $s_2$ as positive training samples, and the other sub-sequences in $s_1$ as negative training samples. We refer readers to Appendix for more details.

In the decoding stage, we begin with setting the size of the dynamic context to 0, and then determine whether to read more context according to the principle defined in definition \ref{def:2}. 

\subsection{Partial Decoding}
Traditional NMT models are usually trained on bilingual corpora containing only complete sentences. However in our context-aware translation model, information units usually are sub-sentences. Intuitively, the discrepancy between the training and the decoding will lead to a problematic translation, if we use the conventional NMT model to translate such information units. 
On the other hand, conventional NMT models rarely do anticipation. Whereas in simultaneous interpreting, human interpreters often have to anticipate the up-coming input and render a constituent at the same time or even before it is uttered by the speaker.  

In our previous work \cite{DBLP:journals/corr/abs-1810-08398}, training a \textit{wait-k} policy slightly differs from the traditional method. When predicting the first target token, we mask the source content behind the $k-th$ token, in order to make the model learn to anticipate. The prediction of other tokens can also be obtained by moving the mask-window token-by-token from position $k+1$ to the end of the line. According to our practical experiments, this training strategy do help the model anticipate correctly most of the time. 
 
Following our previous work, we propose the partial decoding model, a tailored NMT model for translating the IUs that appear at the beginning of each sentence. As depicted in Figure \ref{fig:partial}, in the training stage, we mask the second sub-sentence both in the source and target side. While translating the first sub-sentence, the model learns to anticipate the content after the comma, and produces a temporary translation that can be further completed with more source context. Clearly, this method relies on the associated sub-sentence pairs in the training data (black text in Figure \ref{fig:partial}). In this paper, we propose an automatic method to acquire such sub-sentence pairs.
\begin{figure*}[t!]
\centering
\includegraphics[width=\linewidth]{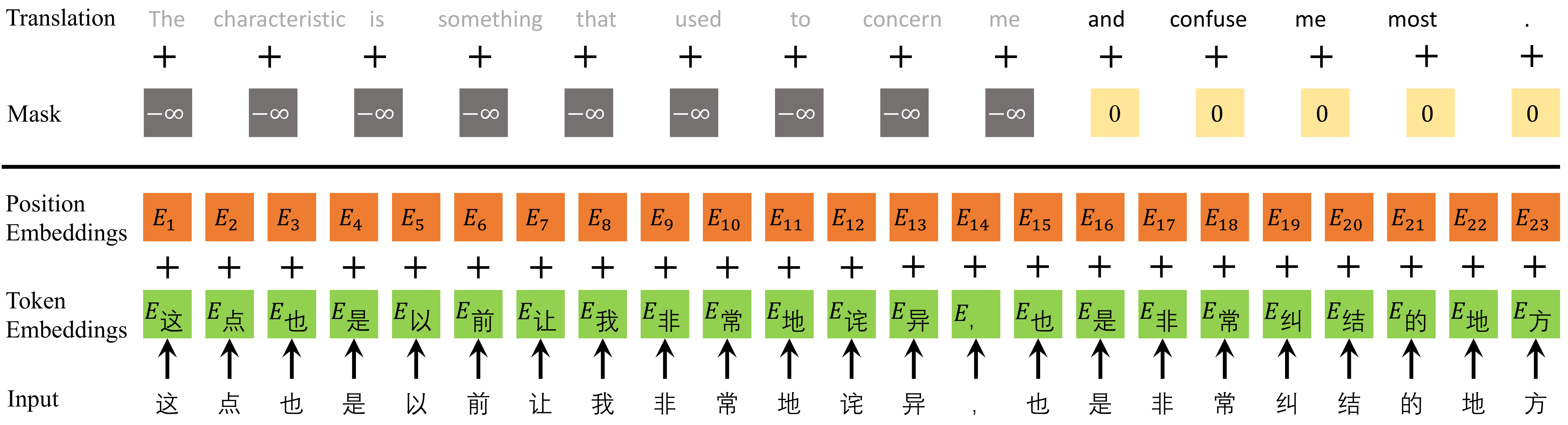}
\caption{Source and target representation for training incremental decoding model. We do not mask the source input, but mask the target sequence aligned to the first sub-sentence. }
\label{fig:incremental}
\end{figure*}
\begin{myDef}
Given a source sentence $X=(x_1,...,x_i,...,x_n)$ with $n$ tokens, a target sentence $Y=(y_1,...,y_j,...,y_m)$ with $m$ tokens, and a word alignment set $A=(a_1,...,a_t,...,a_k)$ where each alignment $a_t=<x_a,y_b>$ is a tuple indicating a word alignment existed between the source token $x_a$ and target token $y_b$, a \textbf{sub-sentence pair} $<(x_1,...,x_i), (y_1,...,y_j)>$ holds if satisfying the following conditions:
\begin{eqnarray}
 <x_i,y_j>\  \in A \\
   \textbf{ if } 1\leqslant a \leqslant i \textbf{ and } b > j \ , \ \textbf{then} <x_a,y_b>\  \notin A  \\
  \textbf{ if } a<i \textbf{ and } 1\leqslant b \leqslant j \ , \ \textbf{then} <x_a,y_b> \ \notin A 
\end{eqnarray}
\label{def:3}
\end{myDef}
To acquire the word alignment, we run the open source toolkit \textit{fast\_align} \footnote{\url{https://github.com/clab/fast\_align}}, and use a variety of standard symmetrization heuristics to generate the alignment matrix. In the training stage, we perform training by firstly tuning the model on a normal bilingual corpus, and then fine-tune the model on a special training corpus containing sub-sentence pairs. 
\subsection{Context-aware Decoding}
For IUs that have one preceding sub-sentence, the context-aware decoding model is applied to translate them based on the pre-generated translations. The requirements of this model are obvious:
\begin{itemize}
\item The model is required to exploit more context to continue the translation.
\item The model is required to generate the coherent translation given partial pre-generated translations.

\end{itemize}

Intuitively, the above requirements can be easily satisfied using a force decoding strategy. For example, when translating the second sub-sentence in ``这点也是以前让我非常地诧异，也是非常纠结的地方'',
given the already-produced translation of the first sub-sentence ``\textit{It also surprised me very much before .}'', the model finishes the translation by adding ``\textit{It's also a very {\color{red}surprising} , tangled place .}''. Clearly, translation is not that accurate and fluent with the redundant constituent ``surprising''. We ascribe this to the discrepancy between training and decoding. In the training stage, the model learns to predict the translation based on the full source sentence. In the decoding stage, the source contexts for translating the first-subsentence and the second-subsentence are different. Forcing the model to generate identical translation of the first sub-sentence is very likely to cause under-translation or over-translation.  

To produce more adequate and coherent translation, we make the following refinements:
\begin{itemize}
\item During training, we force the model to focus on learning how to continue the translation without over-translation and under-translation.  
\item During decoding, we discard a few previously generated translations, in order to make more fluent translations.
\end{itemize}
As shown in Figure \ref{fig:incremental}, during training, we do not mask the source input, instead we mask the target sequence aligned to the first sub-sentence. This strategy will force the model to learn to complete the half-way done translation, rather than to concentrate on generating a translation of the full sentence. 

\begin{figure}[t!]
\centering
\includegraphics[width=\linewidth]{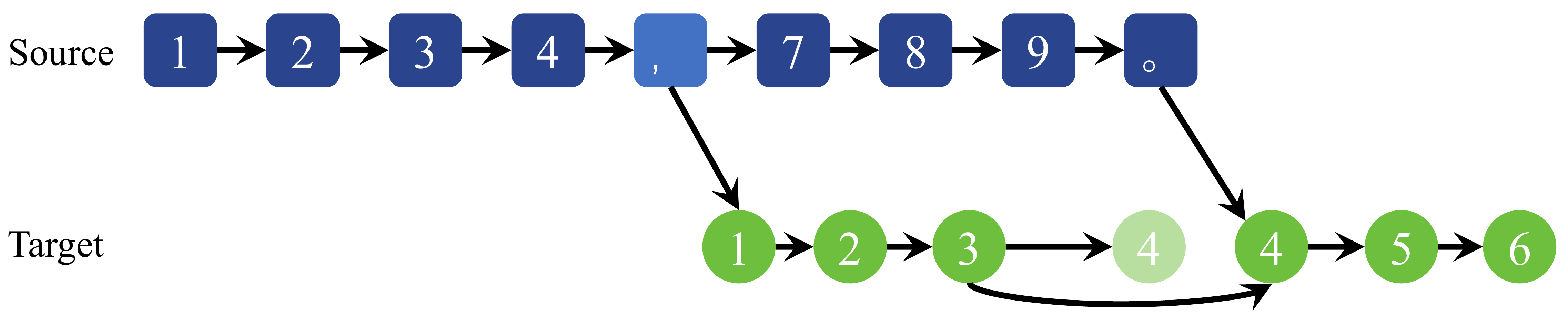}
\caption{In the decoding stage, the context-aware decoding model will discard the last $k$ tokens (in this example, $k=1$) from the generated partial translation to produce a fluent translation. }
\label{fig:incremental2}
\end{figure}

Moreover, in the decoding stage, as shown in Figure \ref{fig:incremental2}, we propose to discard the last $k$ tokens from the generated partial translation (at most times, discarding the last token brings promising result). 
Then the context-aware decoding model will complete the rest of the translation.
The motivation is that the translation of the tail of a sub-sentence is largely influenced by the content of the succeeding sub-sentence. By discarding a few tokens from previously generated translation, the model is able to generate a more appropriate translation. 
 In the practical experiment, this slight modification is proved to be effective in generating fluent translation.

\section{Latency Metric: Equilibrium Efficiency}
In the work of \newcite{DBLP:journals/corr/abs-1810-08398} and \newcite{arivazhagan2019monotonic}, they used the average lagging as the metric for evaluating the latency. However, there are two major flaws of this metric:

1) This metric is unsuitable for evaluating the sub-sentence model. Take the sentence in Figure \ref{fig:intro} for example. As the model reads four tokens ``\underline{她说} \ \underline{我} \ \underline{错了} \ \underline{那个}'', and generates six target tokens ``\textit{She said I was wrong ,}'', the lag of the last target token is one negative value ($g(t)=4; t=6; g(t)-(t-1)=-1$) according to its original definition. 

2) This metric is unsuitable for evaluating latency in the scenario of speech-to-speech translation. \newcite{DBLP:journals/corr/abs-1810-08398} considered that the target token generated after the \textit{cut-off} point doesn't cause any lag. However, this assumption is only supported in the speech-to-text scenario. In the speech-to-speech scenario, it is necessary to consider the time for playing the last synthesized speech.

Therefore, we instead propose a novel metric, Equilibrium Efficiency (EE), which measures the efficiency of equilibrium strategy.
\begin{myDef}
Consider a sentence with $n$ subsequences, and let $LX_i$ be the length of $i-th$ source subsequence that emits a target subsequence with $LY_i$ tokens. Then the equilibrium efficiency is: $\frac{1}{\mathcal{S}(n-1)+LY_n}$, where $\mathcal{S}(i)$ is defined as:
\begin{equation}   
     \mathcal{S}(i) =  \max(\mathcal{S}(i-1) + r\times(LY_i - LX_{i+1}), 0)  \\
\end{equation}
and $\mathcal{S}(0)=0$, $r$ is an empirical factor. 
\end{myDef}
In practice, we set $r$ to 0.3 for Chinese-English translation (reading about 200 English tokens in one minute). 
The motivation of EE is that one good model should equilibrate the time for playing the target speech to the time for listening to the speaker. 
Assuming playing one word takes one second, the EE actually measures the latency from the audience hearing the final target word to the speaker finishing the speech. For example, the EE of the sentence in Figure \ref{fig:cnmt} is equal to $\frac{1}{8}$, since the time for playing the sequence ``\textit{She said I was wrong}'' is equilibrated to the time for speaker speaking the second sub-sentence ``那个 \ 叫 \ 什么 \ 什么\ 呃\  妖姬''. 
\section{Evaluation}
We conduct multiple experiments to evaluate the effectiveness of our system in many ways.
\subsection{Data Description}
\subsubsection{NIST Chinese-English}
We use a subset of the data available for NIST OpenMT08 task \footnote{1LDC2002E18, LDC2002L27, LDC2002T01,
LDC2003E07, LDC2003E14, LDC2004T07, LDC2005E83,
LDC2005T06, LDC2005T10, LDC2005T34, LDC2006E24,
LDC2006E26, LDC2006E34, LDC2006E86, LDC2006E92,
LDC2006E93, LDC2004T08(HK News, HK Hansards )}. The parallel training corpus contains approximate 2 million sentence pairs. We choose NIST 2006 (NIST06) dataset as our development
set, and the NIST 2002 (NIST02), 2003 (NIST03), 2004 (NIST04)
2005 (NIST05), and 2008 (NIST08) datasets as our
test sets. 

We will use this dataset to evaluate the performance of our partial decoding and context-aware decoding strategy from the perspective of translation quality and latency.
\begin{table*}[t!]
\begin{center}
\begin{tabular}{l|c|c|c|c|c|c|c}
\hline
Dataset & Talks & Utterances  &Transcription &Translation&Audio &CER(1-best) &CER(lattice)  \\
\hline
Train &174&26,553&796,679&2,292,025&50.57&17.32\%&15.68\%\\
Dev &16& 956&26,059&75,074 & 1.58&15.21\%&13.20\%\\
Test&6& 975& 25,832&70,503&1.46&10.32\%&8.57\%\\
\hline
\end{tabular}
\end{center}
\caption{The summary of our proposed speech translation data. The volume of transcription is counted by characters, the volume of translation is counted by tokens, and the audio duration is counted by hours.}
\label{tbl:data}
\end{table*}

\subsubsection{BSTC Chinese-English}
Recently, we release Baidu Speech Translation Corpus (BSTC) for open research \footnote{\url{http://ai.baidu.com/broad/subordinate?dataset=bstc}}.
This dataset covers speeches in a wide range of domains, including IT, economy, culture, biology, arts, etc. We transcribe the talks carefully, and have professional translators to produce the English translations. This procedure is extremely difficult due to the large number of domain-specific terminologies, speech redundancies and speakers' accents. We expect that this dataset will help the researchers to develop robust NMT models on the speech translation.
In summary, there are many features that distinguish this dataset to the previously related resources: 
\begin{itemize}
\item  Speech irregularities are kept in transcription while omitted in translation (eg. filler words like ``嗯, 呃, 啊'', and unconscious repetitions like ``这个这个呢''), which can be used to evaluate the robustness of the NMT model dealing with spoken language.
\item Each talk's transcription is translated into English by a single translator, and then segmented into bilingual sentence pairs according to the sentence boundaries in the English translations. Therefore, every sentence is translated based on the understanding of the entire talk and is translated faithfully and coherently in global sense.
\item We use the streaming multi-layer truncated attention model (SMLTA) \footnote{\url{http://research.baidu.com/Blog/index-view?id=109}} trained on the large-scale speech corpus ({\em more than 10,000 hours}) and fine-tuned on a number of talk related corpora (\textit{more than 1,000 hours}), to generate the {\em 5-best} automatic recognized text for each acoustic speech. 
\item The test dataset includes interpretations produced by simultaneous interpreters with professional experience. This dataset contributes an essential resource for the comparison between translation and interpretation. 
\end{itemize}

 We randomly extract several talks from the dataset, and divide them into the development and test set. In Table \ref{tbl:data}, 
 we summarize the statistics of our dataset. The average number of utterances per talk is 152.6 in the training set, 59.75 in the dev set, and 162.5 in the test set. 
 
We firstly run the standard Transformer model on the NIST dataset. Then we evaluate the quality of the pre-trained model on our proposed speech translation dataset, and propose effective methods to improve the performance of the baseline. In that the testing data in this dataset contains ASR errors and speech irregularities, it can be used to evaluate the robustness of novel methods.

\subsubsection{Large-scale Chinese-English and English-Chinese} 
In the final deployment, we train our model using a corpus containing approximately 200 million bilingual pairs both in Chinese-English and English-Chinese translation tasks. 
\subsection{Data Preprocess}
To preprocess the Chinese and the English texts, we use an open source Chinese Segmenter \footnote{\url{https://github.com/fxsjy/jieba}} and Moses Tokenizer \footnote{\url{https://github.com/moses-smt/mosesdecoder/blob/master/scripts/tokenizer/tokenizer.perl}}. After tokenization, we convert all English letters into lower case.
 And we use the ``multi-bleu.pl'' \footnote{\url{https://github.com/moses-smt/\\mosesdecoder/blob/master/scripts/generic/multi-bleu.perl}} script to calculate BLEU scores. Except in the large-scale experiments, we conduct byte-pair encoding (BPE) \cite{DBLP:journals/corr/SennrichHB15} for both Chinese and English by setting the vocabulary size to 20K and 18K for Chinese and English, respectively. But in the large-scale experiments, we utilize a joint vocabulary for both Chinese-English and English-Chinese translation tasks, with a vocabulary size of 40K.
\subsection{Model Settings}
We implement our models
 using PaddlePaddle \footnote{\url{https://github.com/paddlepaddle/
paddle}}, an
end-to-end open source deep learning platform developed
by Baidu. It provides a complete suite
of deep learning libraries, tools and service platforms
to make the research and development of
deep learning simple and reliable. For training our dynamic context sequence boundary detector, we use ERNIE \cite{sun2019ernie} as our pre-trained model.
\begin{table*}[t!]
\begin{center}
\begin{tabular}{l|c|c|c|c|c|c}
\hline
Models & NIST02 & NIST03  &NIST04 &NIST05&NIST08&Average  \\
\hline
\textit{baseline} &49.40&49.71&50.03&48.83&44.38&40.39\\ 
\textit{sub-sentence} &45.41& 45.62&46.06&43.63 & 43.11&37.31\\
\hline
\textit{wait-1} &38.37&36.87 &38.17&36.09 &35.31 &30.80 \\
\textit{wait-3} &40.75&39.30 &40.57&38.18 &38.29 &32.85\\
\textit{wait-5} &42.76&41.43 &43.29&40.43 &39.62 &34.59\\
\textit{wait-7} &44.05&42.94 &44.17&42.25 &40.61 &35.67\\
\textit{wait-9} &45.71&44.49 &45.74&43.14 &41.63 &36.78\\
\textit{wait-12} &\textbf{46.67}&45.63 &46.86&44.59 &42.83 &37.76\\
\textit{wait-15} &46.41&\textbf{46.43} &\textbf{47.38}&\textbf{45.63} &\textbf{43.60} &\textbf{38.24}\\
\hline
\multicolumn{7}{c}{treat the information unit as sub-sentence (\textbf{IU=sub-sentence})} \\
\hline
\textit{+context-aware}&47.79& 48.11& 48.29&46.55&44.57&39.22\\
\textit{+partial decoding}&48.46& 48.51& 48.53&47.05&45.43&39.66\\
\textit{+discard 2 tokens}&48.61& 48.54& 48.68&47.11&45.08&39.67\\
\textit{+discard 3 tokens}&48.62& 48.52& 48.87&47.16&\textbf{45.30}&39.75\\
\textit{+discard 4 tokens}&48.71& 48.69& \textbf{49.10}&\textbf{47.32}&45.11&\textbf{39.82}\\
\textit{+discard 5 tokens}&48.82& \textbf{48.78}& 48.98&47.31&44.48&39.73\\
\textit{+discard 6 tokens}&\textbf{48.94}& 48.70& 48.77&47.21&44.33&39.66\\
\hline
\multicolumn{7}{c}{treat the information unit as segment (\textbf{IU=segment})} \\
\hline
\textit{+discard 1 tokens}&46.89& 45.40& 47.05&45.36&43.06&37.96\\
\textit{+discard 2 tokens}&48.09& 46.98& 48.45&46.50&44.00&39.00\\
\textit{+discard 3 tokens}&48.70& 47.87& 48.85&47.01&{44.48}&39.49\\
\textit{+discard 4 tokens}&48.75& 48.09& \textbf{48.99}&46.86&\textbf{45.07}&{39.63}\\
\textit{+discard 5 tokens}&48.84& 48.37& 48.71&\textbf{46.95}&44.76&39.56\\
\textit{+discard 6 tokens}&\textbf{48.88}& \textbf{48.60}& 48.85&47.17&44.84&\textbf{39.72}\\
\hline
\end{tabular}
\end{center}
\caption{The overall results on NIST Chinese-English translation task.}
\label{tbl:nist1}
\end{table*}

For fair comparison, we implement the following models:
\begin{itemize}
\item \textit{baseline}: A standard Transformer based model with big version of hyper parameters.
\item \textit{sub-sentence}: We split a full sentence into multiple sub-sentences by comma, and translate them using the \textit{baseline} model. To evaluate the translation quality, we concatenate the translation of each sub-sentence into one sentence.
\item \textit{wait-k}: This is our previous work \cite{DBLP:journals/corr/abs-1810-08398}. 
\item \textit{context-aware}: This is our proposed model using context-aware decoding strategy, without fine-tuning on partial decoding model.
\item \textit{partial decoding}: This is our proposed model using partial decoding.
\item \textit{discard $n$ tokens}: The previously generated $n$ tokens are removed to complete the rest of the translation by the context-aware decoding model.
\end{itemize}

\subsection{Experiments}
\subsubsection{NIST Chinese-English}
We firstly conduct our experiments on the NIST Chinese-English translation task.

To validate the effectiveness of our translation model, we run two baseline models, \textit{baseline} and \textit{sub-sentence}. We also compare the translation quality as well as latency of our models with the \textit{wait-k} model. \\

\noindent\textbf{Effectiveness on Translation Quality.} As shown in Table \ref{tbl:nist1}, there is a great deal of difference between the \textit{sub-sentence} and the \textit{baseline} model. On an average the \textit{sub-sentence} shows weaker performance by a 3.08 drop in BLEU score (40.39 $\rightarrow$ 37.31). Similarly, the \textit{wait-k} model also brings an obvious decrease in translation quality, even with the best \textit{wait-15} policy, its performance is still worse than the \textit{baseline} system, with a 2.15 drop, averagely, in BLEU (40.39 $\rightarrow$ 38.24). For a machine translation product, a large degradation in translation quality will largely affect the use experience even if it has low latency. 

\begin{figure}[t]
 \centering
   \centering
   \includegraphics[width=\linewidth]{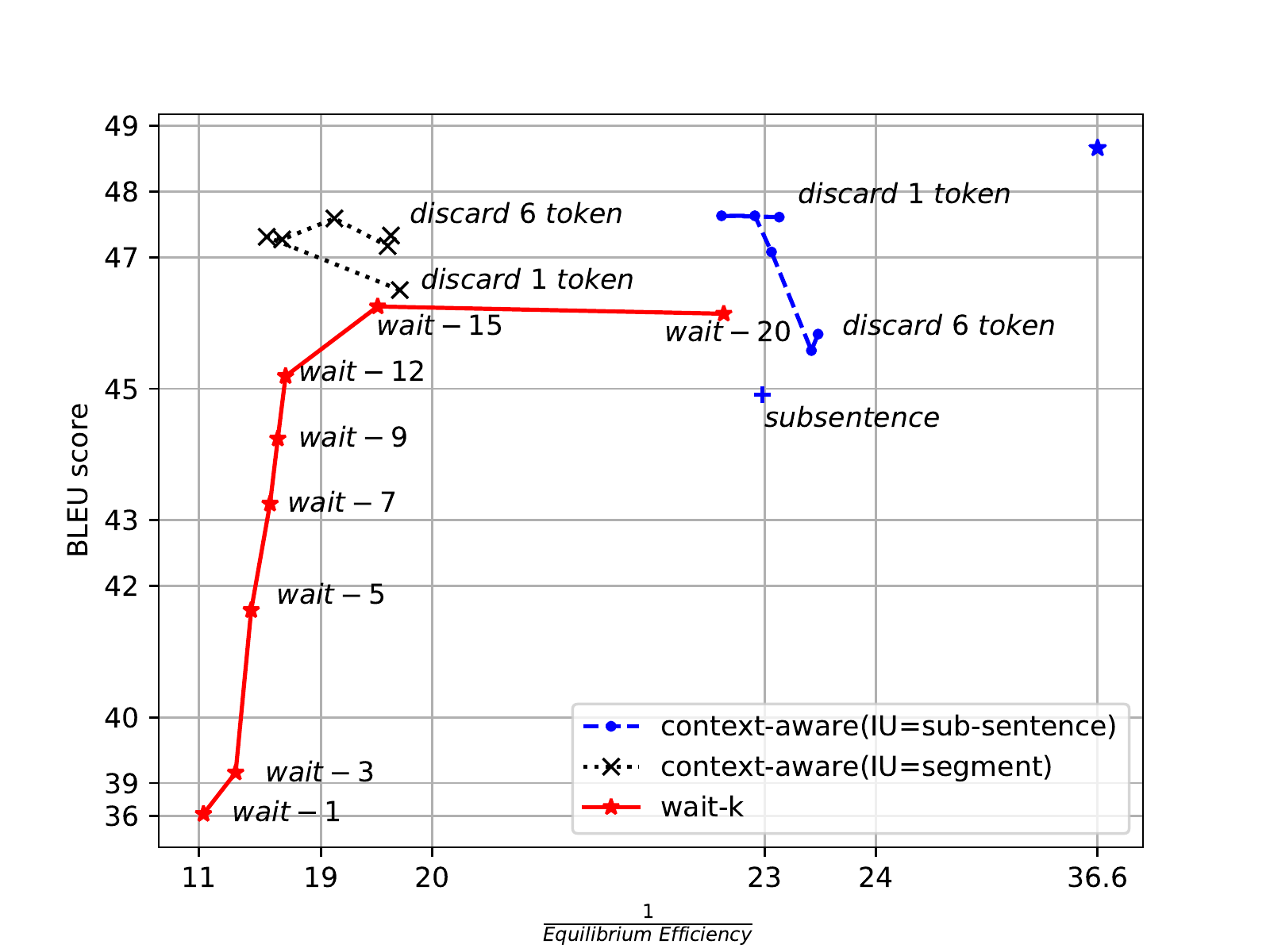}
\caption{We show the latency for our proposed model (Left), and $wait-k$ model (Right). For better understanding, we use the $\frac{1}{Equilibrium \ Efficiency}$ to represent the latency. The $\star$ represents the \textit{baseline} system, and the \textbf{+} represents the \textit{sub-sentence} system.}
\label{fig:ee}
\end{figure}

Unsurprisingly, when treating sub-sentences as IUs, our proposed model significantly improves the translation quality by an average of 2.35 increase in BLEU score (37.31 $\rightarrow$ 39.66), and its performance is slightly lower than the \textit{baseline} system with a 0.73 lower average BLEU score (40.39 $\rightarrow$ 39.66). Moreover, as we allow the model to discard a few previously generated tokens, the performance can be further improved to 39.82 ($+$ 0.16), at a small cost of longer latency (see  Figure \ref{fig:ee}). It is consistent with our intuition that our novel partial decoding strategy can bring stable improvement on each testing dataset. It achieves an average improvement of 0.44 BLEU score (39.22 $\rightarrow$ 39.66) compared to the \textit{context-aware} system in which we do not fine-tune the trained model when using partial decoding strategy. An interesting finding is that our translation model performs better than the \textit{baseline} system on the NIST08 testing set. We analyze the translation results and find that the sentences in NIST08 are extremely long, which affect the standard Transformer to learn better representation \cite{D18-1458}. Using context-aware decoding strategy to generate consistent and coherent translation, our model performs better by focusing on generating translation for relatively shorter sub-sentences.\\  \\

\noindent\textbf{Investigation on Decoding Based on Segment.}
Intuitively, treating one segment as an IU will reduce the latency in waiting for more input to come.
Therefore, we split the testing data into segments according to the principle in Definition \ref{def:3}  (if $x_i$ in Definition \ref{def:3} is a comma, then the data is sub-sentence pair, otherwise it is a segment-pair.) \footnote{Clearly, this generated testing data is unavailable in real product due to the requirement of target translation to extract the segment-pairs. In actual, it is necessary to let the sequence detector making decision upon the segment-level.  }.

As Table \ref{tbl:nist1} shows, although the translation quality of \textit{discard 1 token} based on segment is worse than that based on sub-sentence (37.96 vs. 39.66), the performance can be significantly improved by allowing the model discarding more previously generated tokens. 
Lastly, the \textit{discard 6 tokens} obtains an impressive result, with an average improvement of 1.76 BLEU score (37.96 $\rightarrow$ 39.72). \\

\noindent\textbf{Effects of Discarding Preceding Generated Tokens.}
As mentioned and depicted in Figure \ref{fig:incremental2}, we discard one token in the previously generated translation in our context-aware NMT model. One may be interested in whether discarding more generated translation leads to better translation quality. However, when decoding on the sub-sentence, even the best \textit{discard 4 tokens} model brings no significant improvement (39.66 $\rightarrow$ 39.82) but a slight cost of latency (see in Figure \ref{fig:ee} for visualized latency). While decoding on the segment, even discarding two tokens can bring significant improvement (37.96 $\rightarrow$ 39.00). This finding proves that our partial decoding model is able to generate accurate translation by anticipating the future content. It also indicates that the anticipation based on a larger context presents more robust performance than the aggressive anticipation in the \textit{wait-k} model, as well as in the \textit{segment based decoding} model.\\
\begin{table*}[t!]
\begin{center}
\begin{tabular}{l|c|c|c|c|c}
\hline
Models & Precision (\%) & Recall (\%) & F-score (\%) & Average Latency & Max Latency  \\
\hline
\textit{5-LM} &55.30 & 72.63 & 62.79 & 8.68 & 46 \\
\textit{RNN}&67.61 & 70.35 & 68.95 & 9.79 & 48 \\
\textit{Our model} &\textbf{75.09} & \textbf{81.70}& \textbf{78.26} & 10.49 & 39 \\
\hline
\end{tabular}
\end{center}
\caption{The comparison between our sequence detector and previous work. The latency represents the words requiring to make an explicit decision.}
\label{tbl:sbd}
\end{table*}

\noindent\textbf{Effectiveness on latency.} As latency in simultaneous machine translation is essential and is worth to be intensively investigated, we compare the latency of our models with that of the previous work using our Equilibrium Efficiency metric. As shown in Figure \ref{fig:ee}, we plot the translation quality and $\frac{1}{Equilibrium \ Efficiency}$ on the NIST06 dev set. Clearly, compared to the \textit{baseline} system, our model significantly reduce the time delay while remains a competitive translation quality. When treating segments as IUs, the latency can be further reduced by approximate 20\% (23.13 $\rightarrow$18.65), with a slight decrease in BLEU score (47.61 $\rightarrow$ 47.27).
One interesting finding is that the granularity of information units largely affects both the translation quality and latency. It is clear the decoding based on sub-sentence and based on segment present different performance in two metrics. For the former model, the increase of discarded tokens results in an obvious decrease in translation quality, but no definite improvement in latency. The latter model can benefit from the increasing of discarding tokens both in translation quality and latency.

The latency of the \textit{wait-k} models are competitive, their translation quality, however, is still worse than context-aware model. Improving the translation quality for the \textit{wait-k} will clearly brings a large cost of latency (36.53 $\rightarrow$ 46.14 vs. 10.94 $\rightarrow$ 22.63). Even with a best \textit{k-20} policy, its performance is still worse than most context-aware models.
 More importantly, the intermediately generated target token in the \textit{wait-k} policy is unsuitable for TTS due to the fact that the generated token is often a unit in BPE, typically is an incomplete word. One can certainly wait more target tokens to synthesize the target speech, however, this method will reduce to the \textit{baseline} model. In general, experienced human interpreters lag approximately 5 seconds (15 $\sim$ 25 words) behind the speaker \cite{lee2002ear,lamberger2001text,timarova2015simultaneous}, which indicates that the latency of our model is accessible and practicable ($\frac{1}{Equilibrium \ Efficiency}$ = 25 indicates lagging 25 words).


\subsubsection{Dynamic Context based Information Unit Boundary Detector}

In our context-sensitive model, the dynamic context based information unit boundary detector is essential to determine the IU boundaries in the steaming input. To measure the effectiveness of this model, we compare its precision as well as latency against the traditional language model based methods, a 5-gram language model trained by KenLM toolkit \footnote{\url{https://github.com/kpu/kenlm}}, and an in-house implemented RNN based model. Both of two contrastive models are trained on approximate 2 million monolingual Chinese sentences. As shown in Table \ref{tbl:sbd}, it is clear that our model beats the previous work with an absolute improvement of more than 15 points in term of F-score (62.79 $\rightarrow$ 78.26) and no obvious burden in latency (average latency). This observation indicates that with bidirectional context, the model can learn better representation to help the downstream tasks. In the next experiments, we will evaluate models given testing data with IU boundaries detected by our detector. 
\subsubsection{BSTC Chinese-English}
To our knowledge, almost all of the previous related work on simultaneous translation evaluate their models upon the clean testing data without ASR errors and with explicit sentence boundaries annotated by human translators. Certainly, testing data with real ASR errors and without explicit sentence boundaries is beneficial to evaluate the robustness of translation models. To this end, we perform experiments on our proposed BSTC dataset.
\begin{table*}[t!]
\begin{center}
\begin{tabular}{l|c|c|c|c|c|c}
\hline
\multirow{2}{*}{Models} & \multicolumn{2}{c|}{Clean Input } & \multicolumn{2}{c|}{ASR Input } &\multicolumn{2}{c}{ ASR + Auto IU }  \\
\cline{2-7}
 &Pre-train & Fine-tune & Pre-train & Fine-tune &Pre-train &Fine-tune \\
\hline
\textit{baseline}	&\textbf{15.85}	&	\textbf{21.98}	&	\textbf{14.60}	&	\textbf{19.91}	&	\textbf{14.41}	&	\textbf{17.35}	\\
\textit{sub-sentence}	&14.39	&	18.61	&	13.50	&	16.99	&	13.76	&	16.29	\\

\hline
\textit{wait-3}	&12.23	&	16.74	&	11.62	&	15.59	&	11.75	&	14.68	\\
\textit{wait-5}	&12.84	&	17.70	&	11.96	&	16.23	&	12.25	&	15.45	\\
\textit{wait-7}	&13.34	&	19.32	&	12.67	&	17.41	&	12.55	&	16.08	\\
\textit{wait-9}	&13.92	&	19.77	&	13.05	&	18.29	&	13.12	&	16.49	\\
\textit{wait-12}	&14.35	&	20.15	&	13.34	&	19.07	&	13.48	&	\textbf{17.25}	\\
\textit{wait-15}	&\textbf{14.70}	&	\textbf{21.11}	&	\textbf{13.56}	&	\textbf{19.53}	&	\textbf{13.70}	&	17.21	\\
\hline
\textit{context-aware}&	15.25	&	20.72	&	14.24	&	18.42	&	13.52	&	16.83	\\
\textit{+discard 2 tokens}	&15.26	&	21.07	&	14.35	&	19.17	&	13.73	&	17.02	\\
\textit{+discard 3 tokens}	&15.37	&	21.09	&	14.42	&	19.39	&	14.00	&	17.41	\\
\textit{+discard 4 tokens}	&15.40	&	21.02	&	14.45	&	19.41	&	14.11	&	17.36	\\
\textit{+discard 5 tokens}	&\textbf{15.59}	&	\textbf{21.23}	&	14.72	&	\textbf{19.65}	&	\textbf{14.54}	&	17.37	\\
\textit{+discard 6 tokens}	&15.53	&	21.21	&	\textbf{14.77}	&	19.48	&	14.58	&	\textbf{17.49}	\\
\hline
\end{tabular}
\end{center}
\caption{The overall results on BSTC Chinese-English translation task (Pre-train represents training on the NIST dataset, and fine-tune represents fine-tuning on the BSTC dataset.). Clean input indicates the input is from human annotated transcription, while the ASR input represents the input contains ASR errors. ASR + Auto IU indicates that the sentence boundary as well as sub-sentence is detected by our IU detector. Therefore, this data basically reflects the real environment in practical product. }
\label{tbl:bstc}
\end{table*}

The testing data in BSTC corpus consists of six talks. We firstly employ our ASR model to recognize the acoustic waves into Chinese text, which will be further segmented into small pieces of sub-sentences by our IU detector. To evaluate the contribution of our proposed BSTC dataset, we firstly train all models on the NIST dataset, and then check whether the performance can be further improved by fine-tuning them on the BSTC dataset. 

From the results shown in Table \ref{tbl:bstc}, we conclude the following observations: 
\begin{itemize}
\item  Due to the relatively lower CER in ASR errors (10.32 \%), the distinction between the clean input and the noisy input results in a BLEU score difference smaller than 2 points (15.85 vs. 14.60 for pre-train, and 21.98 vs. 19.91 for fine-tune). 
\item  Despite the small size of the training data in BSTC, fine-tuning on this data is essential to improve the performance of all models. 
\item  In all settings, the best system in \textit{context-aware} model beats the \textit{wait-15} model. 
\item Pre-trained models are not sensitive to errors from Auto IU, while fine-tuned models are. 
\end{itemize}

\begin{table*}[t!]
\begin{center}
\begin{tabular}{l|c|c|c|c}
\hline
\multirow{2}{*}{Models}& \multicolumn{2}{c|}{Translation Reference} &\multicolumn{2}{c}{Interpretation Reference (3-references)} \\
\cline{2-5}
&BLEU &Brevity Penalty & BLEU &Brevity Penalty \\
\hline
\textit{Our Model} & 20.93 & 1.000 & 28.08 & 1.000 \\
\textit{S} & 16.02 & 0.845 & - & -\\
\textit{A} & 16.38 & 0.887 & - & -\\
\textit{B} & 12.08 & 0.893 & - & -\\
\hline
\end{tabular}
\end{center}
\caption{Comparison between machine translation and human interpretation. The interpretation reference consists of a collection of interpretations from S, A and B. Our model is trained on the large-scale corpus.}
\label{tbl:human}
\end{table*}
\begin{table*}[t!]
\begin{center}
\begin{tabular}{l|c|c|c|c|c}
\hline
\multirow{2}{*}{Models}&\multicolumn{4}{c|}{Overall} & \multirow{2}{*}{Missing Translation} \\
\cline{2-5}
 & BAD & OK& GOOD&Acceptability  &   \\
\hline
\textit{Our Model}  &26.09\% & 29.13\% & 44.78\% & \textbf{73.91\%}  & 20\%   \\
S  &36.96\% & 30.87\% & 32.17\% & \textbf{63.04\%}  & 53\%   \\
A & 26.96\% & 35.65\% &37.39\% & \textbf{73.04\%} & 47\%   \\
B  &52.17\% & 31.74\% & 16.09\% & \textbf{47.83\%}  & 53\%   \\

\hline
\end{tabular}
\end{center}
\caption{Results of human evaluation for interpreting and machine translation. Missing Translation indicates the proportion of missing translation in all translation errors. Notice that inadequate translations are marked as \textbf{BAD} by the human translator. }
\label{tbl:humaneval}
\end{table*}

\subsubsection{Machine Translation vs. Human Interpretation}
Another interesting work is to compare machine translation with human interpretation. 
We request three simultaneous interpreters (S, A and B) with years of interpreting experience ranging from three to seven years, to interpret the talks in BSTC testing dataset, in a mock conference setting \footnote{We provide the conferences video of the talks to the interpreters, because in real conferences interpreters have a good view of speakers from the booth.}. 

We concatenate the translation of each talk into one big sentence, and then evaluate it by BLEU score. From Table \ref{tbl:human}, we find that machine translation beats the human interpreters significantly. Moreover, the length of interpretations are relatively short, and results in a high length penalty provided by the evaluation script.  
The result is unsurprising, because human interpreters often deliberately skip non-primary information to keep a reasonable ear-voice span, which may bring a loss of adequacy and yet a shorter lag time, whereas the machine translation model translates the content adequately. 
We also use human interpreting results as references. As Table \ref{tbl:human} indicates, our model achieves a higher BLEU score, 28.08.
 \begin{figure*}[h]
\centering
\includegraphics[width=\linewidth]{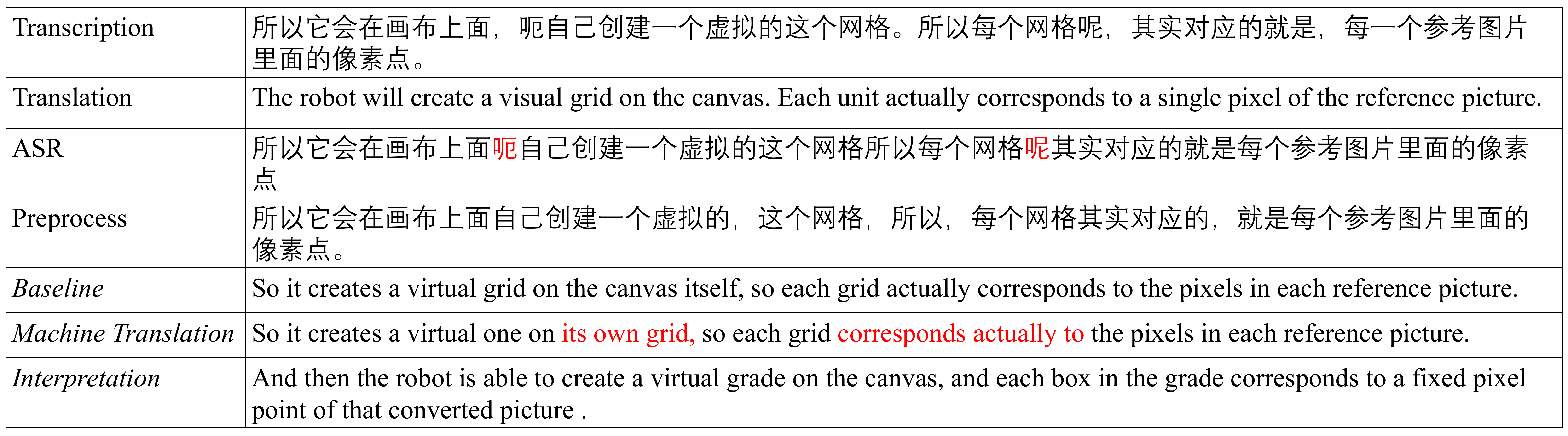}
\caption{This is a representative case that indicates our model can generate coherent translation.}
\label{fig:case}
\end{figure*}
 \begin{table*}[t!]
\begin{center}
\begin{tabular}{l|c|c|c|c|c|c|c}
\hline
\multirow{2}{*}{Task}&\multicolumn{4}{c|}{Overall} & \multicolumn{3}{c}{Error Distribution} \\
\cline{2-8}
 & BAD & OK& GOOD&Acceptability  & Translation&ASR &IU Boundary  \\
\hline

C$\rightarrow$E & 24.29\% & 24.37\% &61.34\% & \textbf{85.71\%} & 13\% &39\% & 48\% \\
E$\rightarrow$C &13.64\% & 14.54\% & 71.82\% & \textbf{86.36\%}  & 15\% & 15\% &70\% \\
\hline
\end{tabular}
\end{center}
\caption{Results of DuTongChuan. C$\rightarrow$E represents Chinese-English translation task, and E$\rightarrow$C represents the English-Chinese translation task.}
\label{tbl:online}
\end{table*}

Furthermore, we ask human translators to evaluate the quality between interpreting and machine translation. 
To evaluate the performance of our final system, we select one Chinese talk as well as one English talk \footnote{\url{https://www.youtube.com/watch?v=RXGNbTx2Wqk}} consisting of about 110 sentences, and have human translators to assess the translation from multiple aspects: adequacy, fluency and correctness.
The detailed measurements are:
\begin{itemize}
\item \textbf{Bad}: Typically, the mark \textbf{Bad} indicates that the translation is incorrect and unacceptable. 
\item \textbf{OK}: If a translation is comprehensible and adequate, but with minor errors such as incorrect function words and less fluent phrases, then it will be marked as \textbf{OK}. 
\item \textbf{Good}: A translation will be marked as \textbf{Good} if it contains no obvious errors.   
\end{itemize}
As shown in Table \ref{tbl:humaneval}, the performance of our model is comparable to the interpreting. 
It is worth mentioning that both automatic and human evaluation criteria are designed for evaluating written translation and have a special emphasis on adequacy and faithfulness. But in simultaneous interpreting, human interpreters routinely omit less-important information to overcome their limitations in working memory. As the last column in Table 6 shows, human interpreters' oral translations have more omissions than machine's and receive lower acceptability. The evaluation results do not mean that machines have exceeded human interpreters in simultaneous interpreting. Instead, it means we need machine translation criteria that suit simultaneous interpreting.
We also find that the BSTC dataset is extremely difficult as the best human interpreter obtains a lower Acceptability 73.04\%. 
 Although the NMT model obtains impressive translation quality, we do not compare the latency of machine translation and human interpreting in this paper, and leave it to the future work. 

\subsubsection{Ablation Study}
To better understand the contribution of our model on generating coherent translation, we select one representative running example for analysis. As the red text in Figure \ref{fig:case} demonstrates that machine translation model generates coherent translation \textit{``its own grid''} for the sub-sentence \textit{``这个网络''}, and \textit{``corresponds actually to''} for the subsequence \textit{``...对应的,就是每个...''}. Compared to the human interpretation, our model presents comparable translation quality. In details, our model treats segments as IUs, and generates translation for each IU consecutively. While the human interpreter splits the entire source text into two sub-sentences, and produces the translation respectively.  
 \subsubsection{Performance of DuTongChuan}
In the final deployment, we train DuTongChuan on the large-scale training corpus. We also utilize techniques to enhance the robustness of the translation model, such as normalization of the speech irregularities, dealing with abnormal ASR errors, and content censorship, etc (see Appendix). We successfully deploy DuTongChuan in the Baidu Create 2019 (Baidu AI Developer Conference) \footnote{\url{https://create.baidu.com/}}. 

As shown in Table \ref{tbl:online}, it is clear that DuTongChuan achieves promising acceptability on both translation tasks (85.71\% for Chinese-English, and 86.36 \% for English-Chinese). We also elaborately analyze the error types in the final translations, and we find that apart from errors occurring in translation and ASR, a majority of errors come from IU boundary detection, which account for nearly a half of errors. In the future, we should concentrate on improving the translation quality by enhancing the robustness of our IU boundary detector. We also evaluate the latency of our model in an End-to-End manner (speech-to-speech), and we find that the target speech slightly lags behind the source speech in less than 3 seconds at most times. The overall performance both on translation quality and latency reveals that DuTongChuan is accessible and practicable in an industrial scenario.

\section{Related Work}
The existing research on speech translation can be divided into two types: the {\em End-to-End} model \cite{duong2016attentional,bansal2017towards,weiss2017sequence,berard2018end,liu2019end} and the {\em cascaded} model. The former approach directly translates the acoustic speech in one language, into text in another language without generating the intermediate transcription for the source language. Depending on the complexity of the translation task as well as the scarce training data, previous literatures explore effective techniques to boost the performance. For example {\em pre-training} \cite{DBLP:journals/corr/abs-1809-01431}, {\em multi-task learning} \cite{duong2016attentional,berard2018end}, {\em attention-passing}, \cite{sperber19tacl}, and \textit{knowledge distillation} \cite{liu2019end} {\em etc.,}. However, the {\em cascaded} model remains the dominant approach and presents superior performance practically, since the ASR and NMT model can be optimized separately training on the large-scale corpus.    

Many studies have proposed to synthesize realistic ASR errors, and augment them with translation training data, to enhance the robustness of the NMT model towards ASR errors \cite{tsvetkov2014augmenting,chen2017mitigating,sperber2017toward}. However, most of these approaches depend on simple  heuristic rules and only evaluate on artificially noisy test set, which do not always reflect the real noises distribution on training and  inference \cite{cheng2018towards,liu2018robust,li2018improving}.

Beyond the research on translation models, there are many research on the other relevant problems, such as sentence boundary detection for realtime speech translation \cite{sridhar2013segmentation,oda2014optimizing,wang2016efficient,bourlon2016simultaneous,zhou2017dynamic}, low-latency simultaneous interpreting \cite{fujita2013simple,grissom2014don,cho2016can,gu2017learning,niehues2018,alinejad2018prediction,press2018you}, automatic punctuation annotation for speech transcription \cite{gravano2009restoring,cho2017nmt}, and discussion about human and machine in simultaneous interpreting \cite{he2016interpretese}. 

Focus on the simultaneous translation task, there are some work referring to the construction of the simultaneous interpreting corpus \cite{tohyama2004ciair,bendazzoli2005approach,shimizu2014collection}. Particularly, \cite{shimizu2014collection} deliver a collection of a simultaneous translation corpus
for comparative analysis on Japanese-English and English-Japanese speech translation. This work analyze the difference between the translation and the interpretations, using the interpretations from human simultaneous interpreters. 

For better generation of coherent translations, \newcite{gong2011cache} propose a memory based approach to capture contextual information to make the statistical translation model generate discourse coherent translations. \newcite{kuang2017cache,tu2018learning,P18-1118} extend similar memory based approach to the NMT framework.
\newcite{wang2017exploiting} present a novel document RNN to learn the representation of the entire text, and treated the external context as the auxiliary context which will be retrieved by the hidden state in the decoder.
 \newcite{tiedemann2017neural} and \newcite{P18-1117} propose to encode global context through extending the current sentence with one preceding adjacent sentence. Notably, the former is conducted on the recurrent based models while the latter is implemented on the Transformer model.
 Recently, we also propose a reinforcement learning strategy to deliberate the translation so that the model can generate more coherent translations \cite{xiong2018modeling}. 
%
%
\section{Conclusion and Future Work}
In this paper, we propose \textbf{DuTongChuan}, a novel context-aware translation model for simultaneous interpreting.  
This model is able to constantly read streaming text from the ASR model, and simultaneously determine the boundaries of information units one after another. The detected IU is then translated into a fluent translation with two simple yet effective decoding strategies: partial decoding and context-aware decoding.
We also release a novel speech translation corpus, BSTC, to boost the research on robust speech translation task.

With elaborate comparison, our model obtains superior translation quality against the \textit{wait-k} model, but also presents competitive performance in latency. Assessment from human translators reveals that our system achieves promising translation quality (85.71\% for Chinese-English, and 86.36\% for English-Chinese), specially in the sense of surprisingly good discourse coherence. Our system also presents superior performance in latency (delayed in less 3 seconds at most times) in a speech-to-speech simultaneous translation. We also deploy our simultaneous machine translation model in our AI platform, and welcome the other users to enjoy it. 

In the future, we will conduct research on novel method to evaluate the interpreting. 
\section{Acknowledgement}
We thank Ying Chen for improving the written of this paper. We thank Yutao Qu for developing partial modules of DuTongChuan.  
We thank colleagues in Baidu for their efforts on construction of the BSTC. They are Zhi Li, Ying Chen, Xuesi Song, Na Chen, Qingfei Li, Xin Hua, Can Jin, Lin Su, Lin Gao, Yang Luo, Xing Wan, Qiaoqiao She, Jingxuan Zhao, Can Jin, Wei Jin, Xiao Yang, Shuo Liu, Yang Zhang, Jing Ma, Junjin Zhao, Yan Xie, Minyang Zhang, Niandong Du, etc.  

We also thank \textit{tndao.com}\footnote{\url{http://www.tndao.com/about-tndao}} and \textit{zaojiu.com}\footnote{\url{https://www.zaojiu.com/}} for contributing their speech corpora.

\bibliography{tacl2018}

\begin{thebibliography}{52}
\expandafter\ifx\csname natexlab\endcsname\relax\def\natexlab#1{#1}\fi

\bibitem[{Alinejad et~al.(2018)Alinejad, Siahbani, and
  Sarkar}]{alinejad2018prediction}
Ashkan Alinejad, Maryam Siahbani, and Anoop Sarkar. 2018.
\newblock Prediction improves simultaneous neural machine translation.
\newblock In \emph{Proceedings of the 2018 Conference on Empirical Methods in
  Natural Language Processing}, pages 3022--3027.

\bibitem[{Arivazhagan et~al.(2019)Arivazhagan, Cherry, Macherey, Chiu, Yavuz,
  Pang, Li, and Raffel}]{arivazhagan2019monotonic}
Naveen Arivazhagan, Colin Cherry, Wolfgang Macherey, Chung-Cheng Chiu, Semih
  Yavuz, Ruoming Pang, Wei Li, and Colin Raffel. 2019.
\newblock Monotonic infinite lookback attention for simultaneous machine
  translation.
\newblock \emph{arXiv preprint arXiv:1906.05218}.

\bibitem[{Bansal et~al.(2018)Bansal, Kamper, Livescu, Lopez, and
  Goldwater}]{DBLP:journals/corr/abs-1809-01431}
Sameer Bansal, Herman Kamper, Karen Livescu, Adam Lopez, and Sharon Goldwater.
  2018.
\newblock \href {http://arxiv.org/abs/1809.01431} {Pre-training on
  high-resource speech recognition improves low-resource speech-to-text
  translation}.
\newblock \emph{CoRR}, abs/1809.01431.

\bibitem[{Bansal et~al.(2017)Bansal, Kamper, Lopez, and
  Goldwater}]{bansal2017towards}
Sameer Bansal, Herman Kamper, Adam Lopez, and Sharon Goldwater. 2017.
\newblock Towards speech-to-text translation without speech recognition.
\newblock In \emph{EACL 2017}.

\bibitem[{Bendazzoli and Sandrelli(2005)}]{bendazzoli2005approach}
Claudio Bendazzoli and Annalisa Sandrelli. 2005.
\newblock An approach to corpus-based interpreting studies: developing epic
  (european parliament interpreting corpus).
\newblock In \emph{Proceedings of the EU-High-Level Scientific Conference
  Series MuTra 2005—Challenges of Multidimensional Translation}.

\bibitem[{B{\'e}rard et~al.(2018)B{\'e}rard, Besacier, Kocabiyikoglu, and
  Pietquin}]{berard2018end}
Alexandre B{\'e}rard, Laurent Besacier, Ali~Can Kocabiyikoglu, and Olivier
  Pietquin. 2018.
\newblock End-to-end automatic speech translation of audiobooks.
\newblock In \emph{ICASSP 2018}.

\bibitem[{Bourlon et~al.(2016)Bourlon, Chu, Nakazawa, and
  Kurohashi}]{bourlon2016simultaneous}
Antoine Bourlon, Chenhui Chu, Toshiaki Nakazawa, and Sadao Kurohashi. 2016.
\newblock Simultaneous sentence boundary detection and alignment with
  pivot-based machine translation generated lexicons.
\newblock In \emph{LREC}.

\bibitem[{Chen et~al.(2017)Chen, Hsu, Huang, and Lee}]{chen2017mitigating}
Pin-Jung Chen, I-Hung Hsu, Yi-Yao Huang, and Hung-Yi Lee. 2017.
\newblock Mitigating the impact of speech recognition errors on chatbot using
  sequence-to-sequence model.
\newblock In \emph{ASRU}, pages 497--503. IEEE.

\bibitem[{{Cheng} et~al.(2018){Cheng}, {Tu}, {Meng}, {Zhai}, and
  {Liu}}]{cheng2018towards}
Yong {Cheng}, Zhaopeng {Tu}, Fandong {Meng}, Junjie {Zhai}, and Yang {Liu}.
  2018.
\newblock Towards robust neural machine translation.
\newblock \emph{In Proc. ACL}, 1:1756--1766.

\bibitem[{Cho et~al.(2017)Cho, Niehues, and Waibel}]{cho2017nmt}
Eunah Cho, Jan Niehues, and Alex Waibel. 2017.
\newblock Nmt-based segmentation and punctuation insertion for real-time spoken
  language translation.
\newblock In \emph{INTERSPEECH}, pages 2645--2649.

\bibitem[{Cho and Esipova(2016)}]{cho2016can}
Kyunghyun Cho and Masha Esipova. 2016.
\newblock Can neural machine translation do simultaneous translation?
\newblock \emph{arXiv preprint arXiv:1606.02012}.

\bibitem[{Devlin et~al.(2019)Devlin, Chang, Lee, and
  Toutanova}]{devlin2019bert}
Jacob Devlin, Ming-Wei Chang, Kenton Lee, and Kristina Toutanova. 2019.
\newblock Bert: Pre-training of deep bidirectional transformers for language
  understanding.
\newblock In \emph{Proceedings of the 2019 Conference of the North American
  Chapter of the Association for Computational Linguistics: Human Language
  Technologies, Volume 1 (Long and Short Papers)}, pages 4171--4186.

\bibitem[{Duong et~al.(2016)Duong, Anastasopoulos, Chiang, Bird, and
  Cohn}]{duong2016attentional}
Long Duong, Antonios Anastasopoulos, David Chiang, Steven Bird, and Trevor
  Cohn. 2016.
\newblock An attentional model for speech translation without transcription.
\newblock In \emph{Proceedings of the 2016 Conference of the North American
  Chapter of the Association for Computational Linguistics: Human Language
  Technologies}, pages 949--959.

\bibitem[{Fujita et~al.(2013)Fujita, Neubig, Sakti, Toda, and
  Nakamura}]{fujita2013simple}
Tomoki Fujita, Graham Neubig, Sakriani Sakti, Tomoki Toda, and Satoshi
  Nakamura. 2013.
\newblock Simple, lexicalized choice of translation timing for simultaneous
  speech translation.
\newblock In \emph{INTERSPEECH}, pages 3487--3491.

\bibitem[{Gile(2009)}]{gile2009basic}
Daniel Gile. 2009.
\newblock \emph{Basic concepts and models for interpreter and translator
  training}.
\newblock John Benjamins.

\bibitem[{Gong et~al.(2011)Gong, Zhang, and Zhou}]{gong2011cache}
Zhengxian Gong, Min Zhang, and Guodong Zhou. 2011.
\newblock Cache-based document-level statistical machine translation.
\newblock In \emph{Proceedings of the Conference on Empirical Methods in
  Natural Language Processing}, pages 909--919. Association for Computational
  Linguistics.

\bibitem[{Gravano et~al.(2009)Gravano, Jansche, and
  Bacchiani}]{gravano2009restoring}
Agustin Gravano, Martin Jansche, and Michiel Bacchiani. 2009.
\newblock Restoring punctuation and capitalization in transcribed speech.
\newblock In \emph{2009 IEEE International Conference on Acoustics, Speech and
  Signal Processing}, pages 4741--4744. IEEE.

\bibitem[{Grissom~II et~al.(2014)Grissom~II, He, Boyd-Graber, Morgan, and
  Daum{\'e}~III}]{grissom2014don}
Alvin Grissom~II, He~He, Jordan Boyd-Graber, John Morgan, and Hal
  Daum{\'e}~III. 2014.
\newblock Don’t until the final verb wait: Reinforcement learning for
  simultaneous machine translation.
\newblock In \emph{Proceedings of the 2014 Conference on empirical methods in
  natural language processing (EMNLP)}, pages 1342--1352.

\bibitem[{Gu et~al.(2017)Gu, Neubig, Cho, and Li}]{gu2017learning}
Jiatao Gu, Graham Neubig, Kyunghyun Cho, and Victor~OK Li. 2017.
\newblock Learning to translate in real-time with neural machine translation.
\newblock In \emph{Proceedings of the 15th Conference of the European Chapter
  of the Association for Computational Linguistics: Volume 1, Long Papers},
  pages 1053--1062.

\bibitem[{He et~al.(2016)He, Boyd-Graber, and
  Daum{\'e}~III}]{he2016interpretese}
He~He, Jordan Boyd-Graber, and Hal Daum{\'e}~III. 2016.
\newblock Interpretese vs. translationese: The uniqueness of human strategies
  in simultaneous interpretation.
\newblock In \emph{Proceedings of the 2016 Conference of the North American
  Chapter of the Association for Computational Linguistics: Human Language
  Technologies}, pages 971--976.

\bibitem[{Kuang et~al.(2017)Kuang, Xiong, Luo, and Zhou}]{kuang2017cache}
Shaohui Kuang, Deyi Xiong, Weihua Luo, and Guodong Zhou. 2017.
\newblock Cache-based document-level neural machine translation.
\newblock \emph{arXiv preprint arXiv:1711.11221}.

\bibitem[{Lamberger-Felber(2001)}]{lamberger2001text}
Heike Lamberger-Felber. 2001.
\newblock Text-oriented research into interpreting-examples from a case-study.
\newblock \emph{HERMES-Journal of Language and Communication in Business},
  (26):39--64.

\bibitem[{Lee(2002)}]{lee2002ear}
Tae-Hyung Lee. 2002.
\newblock Ear voice span in english into korean simultaneous interpretation.
\newblock \emph{Meta: Journal des traducteurs/Meta: Translators' Journal},
  47(4):596--606.

\bibitem[{Li et~al.(2018)Li, Xue, Chen, Liu, Feng, and Liu}]{li2018improving}
Xiang Li, Haiyang Xue, Wei Chen, Yang Liu, Yang Feng, and Qun Liu. 2018.
\newblock Improving the robustness of speech translation.
\newblock \emph{arXiv preprint arXiv:1811.00728}.

\bibitem[{Liu et~al.(2018)Liu, Ma, Huang, Xiong, and He}]{liu2018robust}
Hairong Liu, Mingbo Ma, Liang Huang, Hao Xiong, and Zhongjun He. 2018.
\newblock Robust neural machine translation with joint textual and phonetic
  embedding.
\newblock \emph{arXiv preprint arXiv:1810.06729}.

\bibitem[{Liu et~al.(2019)Liu, Xiong, He, Zhang, Wu, Wang, and
  Zong}]{liu2019end}
Yuchen Liu, Hao Xiong, Zhongjun He, Jiajun Zhang, Hua Wu, Haifeng Wang, and
  Chengqing Zong. 2019.
\newblock End-to-end speech translation with knowledge distillation.
\newblock In \emph{Interspeech}.

\bibitem[{Ma et~al.(2019)Ma, Huang, Xiong, Liu, Zhang, He, Liu, Li, and
  Wang}]{DBLP:journals/corr/abs-1810-08398}
Mingbo Ma, Liang Huang, Hao Xiong, Kaibo Liu, Chuanqiang Zhang, Zhongjun He,
  Hairong Liu, Xing Li, and Haifeng Wang. 2019.
\newblock \href {http://arxiv.org/abs/1810.08398} {{STACL:} simultaneous
  translation with integrated anticipation and controllable latency}.
\newblock In \emph{ACL 2019}, volume abs/1810.08398.

\bibitem[{Maruf and Haffari(2018)}]{P18-1118}
Sameen Maruf and Gholamreza Haffari. 2018.
\newblock \href {http://aclweb.org/anthology/P18-1118} {Document context neural
  machine translation with memory networks}.
\newblock In \emph{Proceedings of the 56th Annual Meeting of the Association
  for Computational Linguistics (Volume 1: Long Papers)}, pages 1275--1284.
  Association for Computational Linguistics.

\bibitem[{Niehues et~al.(2018)Niehues, Pham, Ha, Sperber, and
  Waibel}]{niehues2018}
Jan Niehues, Quan Pham, Thanh~Le Ha, Matthias Sperber, and Alex Waibel. 2018.
\newblock Low-latency neural speech translation.
\newblock In \emph{Interspeech 2018}, pages 1293--1297.

\bibitem[{Oda et~al.(2014)Oda, Neubig, Sakti, Toda, and
  Nakamura}]{oda2014optimizing}
Yusuke Oda, Graham Neubig, Sakriani Sakti, Tomoki Toda, and Satoshi Nakamura.
  2014.
\newblock Optimizing segmentation strategies for simultaneous speech
  translation.
\newblock In \emph{Proceedings of the 52nd Annual Meeting of the Association
  for Computational Linguistics (Volume 2: Short Papers)}, volume~2, pages
  551--556.

\bibitem[{Post and Vilar(2018)}]{post2018fast}
Matt Post and David Vilar. 2018.
\newblock Fast lexically constrained decoding with dynamic beam allocation for
  neural machine translation.
\newblock In \emph{Proceedings of the 2018 Conference of the North American
  Chapter of the Association for Computational Linguistics: Human Language
  Technologies, Volume 1 (Long Papers)}, pages 1314--1324.

\bibitem[{Press and Smith(2018)}]{press2018you}
Ofir Press and Noah~A Smith. 2018.
\newblock You may not need attention.
\newblock \emph{arXiv preprint arXiv:1810.13409}.

\bibitem[{Roderick(1998)}]{roderick1998conference}
Jones Roderick. 1998.
\newblock Conference interpreting explained.--manchester: St.

\bibitem[{Sennrich et~al.(2016)Sennrich, Haddow, and
  Birch}]{DBLP:journals/corr/SennrichHB15}
Rico Sennrich, Barry Haddow, and Alexandra Birch. 2016.
\newblock Neural machine translation of rare words with subword units.
\newblock In \emph{Proceedings of the 54th Annual Meeting of the Association
  for Computational Linguistics (Volume 1: Long Papers)}, volume~1, pages
  1715--1725.

\bibitem[{Shimizu et~al.(2014)Shimizu, Neubig, Sakti, Toda, and
  Nakamura}]{shimizu2014collection}
Hiroaki Shimizu, Graham Neubig, Sakriani Sakti, Tomoki Toda, and Satoshi
  Nakamura. 2014.
\newblock Collection of a simultaneous translation corpus for comparative
  analysis.
\newblock In \emph{LREC}, pages 670--673. Citeseer.

\bibitem[{Sperber et~al.(2019)Sperber, Neubig, Niehues, and
  Waibel}]{sperber19tacl}
Matthias Sperber, Graham Neubig, Jan Niehues, and Alex Waibel. 2019.
\newblock \href {https://arxiv.org/abs/1904.07209} {Attention-passing models
  for robust and data-efficient end-to-end speech translation}.
\newblock In \emph{Transactions of the Association for Computational
  Linguistics}.

\bibitem[{Sperber et~al.(2017)Sperber, Niehues, and Waibel}]{sperber2017toward}
Matthias Sperber, Jan Niehues, and Alex Waibel. 2017.
\newblock Toward robust neural machine translation for noisy input sequences.
\newblock In \emph{IWSLT}.

\bibitem[{Sridhar et~al.(2013)Sridhar, Chen, Bangalore, Ljolje, and
  Chengalvarayan}]{sridhar2013segmentation}
Vivek Kumar~Rangarajan Sridhar, John Chen, Srinivas Bangalore, Andrej Ljolje,
  and Rathinavelu Chengalvarayan. 2013.
\newblock Segmentation strategies for streaming speech translation.
\newblock In \emph{Proceedings of the 2013 Conference of the North American
  Chapter of the Association for Computational Linguistics: Human Language
  Technologies}, pages 230--238.

\bibitem[{Sun et~al.(2019)Sun, Wang, Li, Feng, Chen, Zhang, Tian, Zhu, Tian,
  and Wu}]{sun2019ernie}
Yu~Sun, Shuohuan Wang, Yukun Li, Shikun Feng, Xuyi Chen, Han Zhang, Xin Tian,
  Danxiang Zhu, Hao Tian, and Hua Wu. 2019.
\newblock Ernie: Enhanced representation through knowledge integration.
\newblock \emph{arXiv preprint arXiv:1904.09223}.

\bibitem[{Tang et~al.(2018)Tang, M{\"u}ller, Rios, and Sennrich}]{D18-1458}
Gongbo Tang, Mathias M{\"u}ller, Annette Rios, and Rico Sennrich. 2018.
\newblock \href {http://aclweb.org/anthology/D18-1458} {Why self-attention? a
  targeted evaluation of neural machine translation architectures}.
\newblock In \emph{Proceedings of the 2018 Conference on Empirical Methods in
  Natural Language Processing}, pages 4263--4272. Association for Computational
  Linguistics.

\bibitem[{Tiedemann and Scherrer(2017)}]{tiedemann2017neural}
J{\"o}rg Tiedemann and Yves Scherrer. 2017.
\newblock Neural machine translation with extended context.
\newblock In \emph{Proceedings of the Third Workshop on Discourse in Machine
  Translation}, pages 82--92.

\bibitem[{Timarov{\'a} et~al.(2015)Timarov{\'a}, {\v{C}}e{\v{n}}kov{\'a},
  Meylaerts, Hertog, Szmalec, and Duyck}]{timarova2015simultaneous}
{\v{S}}{\'a}rka Timarov{\'a}, Ivana {\v{C}}e{\v{n}}kov{\'a}, Reine Meylaerts,
  Erik Hertog, Arnaud Szmalec, and Wouter Duyck. 2015.
\newblock Simultaneous interpreting and working memory capacity.
\newblock \emph{Psycholinguistic and cognitive inquiries into translation and
  interpreting}, pages 101--126.

\bibitem[{Tohyama et~al.(2004)Tohyama, Matsubara, Ryu, Kawaguch, and
  Inagaki}]{tohyama2004ciair}
Hitomi Tohyama, Shigeki Matsubara, Koichiro Ryu, N~Kawaguch, and Yasuyoshi
  Inagaki. 2004.
\newblock Ciair simultaneous interpretation corpus.
\newblock In \emph{Proc. Oriental COCOSDA}.

\bibitem[{Tsvetkov et~al.(2014)Tsvetkov, Metze, and
  Dyer}]{tsvetkov2014augmenting}
Yulia Tsvetkov, Florian Metze, and Chris Dyer. 2014.
\newblock Augmenting translation models with simulated acoustic confusions for
  improved spoken language translation.
\newblock In \emph{In Proc. EACL}, pages 616--625.

\bibitem[{Tu et~al.(2018)Tu, Liu, Shi, and Zhang}]{tu2018learning}
Zhaopeng Tu, Yang Liu, Shuming Shi, and Tong Zhang. 2018.
\newblock Learning to remember translation history with a continuous cache.
\newblock \emph{Transactions of the Association of Computational Linguistics},
  6:407--420.

\bibitem[{Voita et~al.(2018)Voita, Serdyukov, Sennrich, and Titov}]{P18-1117}
Elena Voita, Pavel Serdyukov, Rico Sennrich, and Ivan Titov. 2018.
\newblock \href {http://aclweb.org/anthology/P18-1117} {Context-aware neural
  machine translation learns anaphora resolution}.
\newblock In \emph{Proceedings of the 56th Annual Meeting of the Association
  for Computational Linguistics (Volume 1: Long Papers)}, pages 1264--1274.
  Association for Computational Linguistics.

\bibitem[{Wang et~al.(2017)Wang, Tu, Way, and Liu}]{wang2017exploiting}
Longyue Wang, Zhaopeng Tu, Andy Way, and Qun Liu. 2017.
\newblock Exploiting cross-sentence context for neural machine translation.
\newblock In \emph{Conference on Empirical Methods in Natural Language
  Processing}.

\bibitem[{Wang et~al.(2016)Wang, Finch, Utiyama, and
  Sumita}]{wang2016efficient}
Xiaolin Wang, Andrew Finch, Masao Utiyama, and Eiichiro Sumita. 2016.
\newblock An efficient and effective online sentence segmenter for simultaneous
  interpretation.
\newblock In \emph{Proceedings of the 3rd Workshop on Asian Translation
  (WAT2016)}, pages 139--148.

\bibitem[{Weiss et~al.(2017)Weiss, Chorowski, Jaitly, Wu, and
  Chen}]{weiss2017sequence}
Ron~J Weiss, Jan Chorowski, Navdeep Jaitly, Yonghui Wu, and Zhifeng Chen. 2017.
\newblock Sequence-to-sequence models can directly translate foreign speech.
\newblock \emph{arXiv preprint arXiv:1703.08581}.

\bibitem[{Xiong et~al.(2019)Xiong, He, Wu, and Wang}]{xiong2018modeling}
Hao Xiong, Zhongjun He, Hua Wu, and Haifeng Wang. 2019.
\newblock Modeling coherence for discourse neural machine translation.
\newblock In \emph{AAAI}.

\bibitem[{Yang et~al.(2019)Yang, Dai, Yang, Carbonell, Salakhutdinov, and
  Le}]{yang2019xlnet}
Zhilin Yang, Zihang Dai, Yiming Yang, Jaime Carbonell, Ruslan Salakhutdinov,
  and Quoc~V Le. 2019.
\newblock Xlnet: Generalized autoregressive pretraining for language
  understanding.
\newblock \emph{arXiv preprint arXiv:1906.08237}.

\bibitem[{Zhou et~al.(2017)Zhou, Wang, and Aw}]{zhou2017dynamic}
Nina Zhou, Xuancong Wang, and AiTi Aw. 2017.
\newblock Dynamic boundary detection for speech translation.
\newblock In \emph{2017 Asia-Pacific Signal and Information Processing
  Association Annual Summit and Conference (APSIPA ASC)}, pages 651--656. IEEE.

\end{thebibliography}
\bibliographystyle{acl_natbib}
\clearpage
\appendix
 
\section{Training Samples for Information Unit Detector}
For example, for a sentence ``她说我错了，那个叫什么什么呃妖姬。'', there are some representative training samples:
\begin{figure}[h!]
\centering
\includegraphics[width=\linewidth]{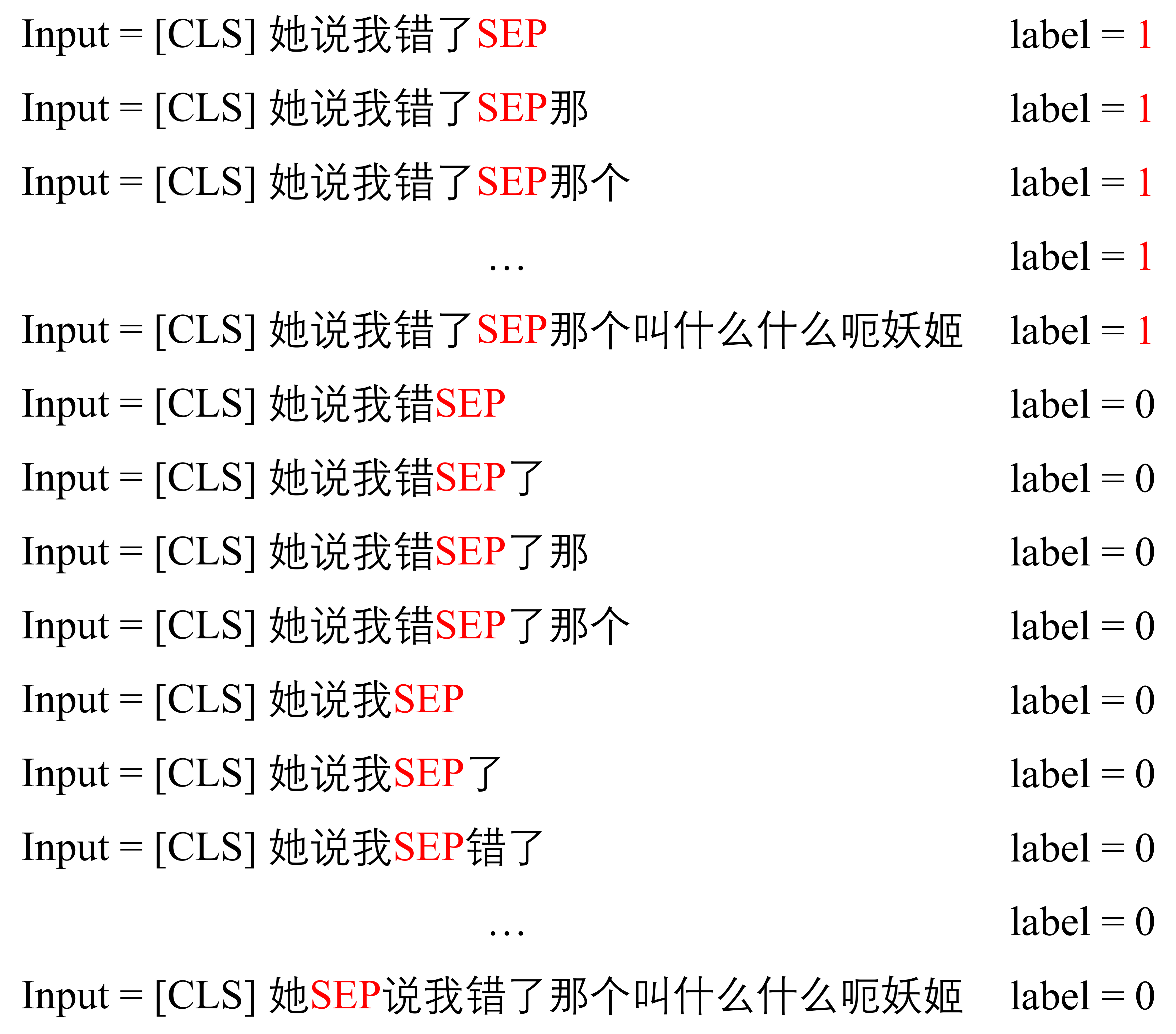}

\label{fig:positivecase}
\end{figure}
\section{Techniques for Robust Translation}
To develop an industrial simultaneous machine translation system, it is necessary to deal with problems that affect the translation quality in practice such as large number of speech irregularities, ASR errors, and topics that allude to violence, religion, sex and politics.
\subsection{Speech Irregularities Normalization}
In the real talk, the speaker tends to express his opinion using irregularities rather than regular written language utilized to train prevalent machine translation relevant models. 
For example, as depicted in Figure \ref{fig:intro}, the spoken language in the real talk often contains unconscious repetitions (i.e., ``什么(sh\'enm\={e})\ 什么(sh\'enm\={e})), and filler words (``呃'', ``啊''), which will inevitably affects the downstream models, especially the NMT model. 
The discrepancy between training and decoding is not only existed in the corpus, but also occurs due to the error propagation from ASR model (e.g. recognize the ``饿 (\`e)'' into filler word ``呃 (\`e) '' erroneously), which is related to the field of robust speech NMT research. 

In the study of robust speech translation, there are many methods can be applied to alleviate the discrepancy mostly arising from the ASR errors such as disfluency detection, fine-tuning on the noisy training data \cite{tsvetkov2014augmenting,chen2017mitigating}, complex lattice input \cite{sperber2017toward}, etc. For spoken language normalization, it is mostly related to the work of sentence simplification. However, the traditional methods for sentence simplification rely large-scale training corpus and will enhance the model complexity by incorporating an End-to-End model to transform the original input. 

In our system, to resolve problems both on speech irregularities and ASR errors, we propose a simple rule heuristic method to normalize both spoken language and ASR errors, mostly focus on removing noisy inputs, including filler words, unconscious repetitions, and ASR error that is easy to be detected.   
Although faithfulness and adequacy is essential in the period of the simultaneous interpreting, however, in a conference, users can understand the majority of the content by discarding some unimportant words. 

\subsubsection{Unconscious Repetitions} To remove unconscious repetitions, the problem can be formulated as the Longest Continuous Substring (LCS) problem, which can be solved by an efficient \textit{suffix-array} based algorithm in $O(n\log n)$ time complexity empirically.
Unfortunately, this simple solution is problematic in some cases. For example, ``他\ 必须\ 分成\ 很多\ 个\ 小格\ ，\ 一个\ 小格\ 一个\ 小格\ 完成'', in this case, the unconscious repetitions ``一个\ 小格\ 一个\ 小格'' can not be normalized to ``一个\ 小格''. To resolve this drawback, we collect unconscious repetitions appearing more than 5 times in a large-scale corpus consisting of written expressions, resulting in a \textit{white list} containing more than 7,000 unconscious repetitions. In practice, we will firstly retrieve this \textit{white list} and prevent the candidates existed in it from being normalized. 

\subsubsection{Removing ASR Errors} 
According to our previous study, many ASR errors are caused by disambiguating homophone. In some cases, such error will lead to serious problem. For example, both ``食油 (cooking oil)'' and ``石油 (oil)'' have similar Chinese phonetic alphabet (sh\'i y\'ou), but with distinct semantics. The simplest method to resolve this problem is to enhance the ASR model by utilizing a domain-specific language model to generate the correct sequence. However, this method requires an insatiably difficult requirement, a customized ASR model. To reduce the cost of deploying a customized ASR model, as well as to alleviate the propagation of ASR errors, we propose a language model based identifier to remove the abnormal contents.
\begin{myDef}
For a given sequence $X_{1,t}=(x_1,...,x_t)$, if the value of $\frac{p(x_1,...,x_t)}{p(x_1,...,x_{t-1})}$ is lower than a threshold $\xi$, then we denote the token $x_t$ as an \textbf{abnormal content}.
\label{def:5}
\end{myDef}
In the above definition, the value of $p(x_1,...,x_t)$ and $p(x_1,...,x_{t-1})$ can be efficiently computed by a language model. In our final system, we firstly train a language model on the domain-specific monolingual corpus, and then identify the abnormal content before the context-aware translation model. For the detected abnormal content, we simply discard it rather than finding an alternative, which will lead to additional errors potentially. Actually, human interpreters often routinely omit source content due to the limited memory.
\subsection{Constrained Decoding and Content Censorship}
For an industrial product, it is extremely important to control the content that will be presented to the audience. Additionally, it is also important to make a consistent translation for the domain-specific entities and terminologies. This two demands lead to two associate problems: \textit{content censorship} and \textit{constrained decoding}, where the former aims to avoid producing some translation while the latter has the opposite target, generating pre-specified translation. 

Recently, \newcite{post2018fast} proposed a Dynamic Beam Allocation (DBA) strategy, a beam search algorithm that forces the inclusion of pre-specified words and phrases in the output. In the DBA strategy, there are many manually annotated constraints, to force the beam search generating the pre-specified translation. To satisfy the requirement of content censorship, we extend this algorithm to prevent the model from generating the pre-specified forbidden content, a collection that contains words and phrases alluding to violence, religion, sex and politics. 
Specially, during the beam search, we punish the candidate beam that matches a constraint of pre-specified forbidden content, to prevent it from being selected as the final translation.
\end{CJK*}

\end{document}